\documentclass[10pt]{article} %
\usepackage[accepted]{tmlr}

\usepackage{hyperref}
\usepackage{url}
\usepackage{amsmath,amsfonts,bm}
\usepackage{comment}
\usepackage[super]{nth}
\usepackage{tikz}
\usepackage{tabularx}
\usepackage{soul}
\usepackage{multirow}
\usepackage{colortbl}
\usepackage{algorithm}
\usepackage{algpseudocode}
\usepackage{dsfont}
\usepackage[dvipsnames,table]{xcolor}
\usepackage{cleveref}
\usepackage{booktabs}
\usepackage{wrapfig}
\usepackage{amsthm}
\usepackage{nicefrac}

\newcommand{\ie}{i.e.}
\newcommand{\eg}{e.g.}
\newcommand{\vs}{vs.}

\newcolumntype{Y}{>{\centering\arraybackslash}X} %
\newcolumntype{C}[1]{>{\centering\arraybackslash}p{#1}} %
\newcolumntype{L}[1]{>{\raggedright\arraybackslash}p{#1}} %

\newcommand{\changed}[1]{#1}

\newcommand{\backbone}[1]{f_{#1}}
\newcommand{\classifier}[1]{h_{#1}}
\newcommand{\mine}{\mathcal{M}}
\newcommand{\model}{\classifier{\varphi} \circ \backbone{\theta}}
\newcommand{\original}{\classifier{\varphi_o} \circ \backbone{\theta_o}}
\newcommand{\unlearned}{\classifier{\varphi_u} \circ \backbone{\theta_u}}
\newcommand{\retrained}{\classifier{\varphi_r} \circ \backbone{\theta_r}}
\newcommand{\training}{\mathcal{T}}

\newcommand{\p}{P}
\newcommand{\reweighting}{\omega}
\newcommand{\optim}{\underset{\varphi, \theta}{\arg\min}}
\newcommand{\sampling}{(\image, \target)\sim\reweighting(\retainds)}
\newcommand{\retainingterm}{\mathcal{L}_r(\reweighting(\retainds); \varphi, \theta)} 
\newcommand{\unlearningterm}{\mine(\forgetds;\psi, \theta)}
\newcommand{\alignmentterm}{\mathcal{L}_{c}(\reweighting(\retainds); \psi, \theta, \theta_o)}

\newcommand{\dataset}{\mathcal{D}}
\newcommand{\trainds}{\dataset_{\mathit{tr}}}
\newcommand{\testds}{\dataset_{\mathit{te}}}

\newcommand{\retainds}{\dataset_{\mathit{r}}}
\newcommand{\forgetds}{\dataset_{\mathit{f}}}
\newcommand{\image}{\mathbf{x}}
\newcommand{\target}{y}
\newcommand{\attribute}{a}
\newcommand{\groups}{G}
\newcommand{\targets}{Y}
\newcommand{\attributes}{A}

\newtheorem{proposition}{Proposition}
\newcommand{\entropy}{\mathcal{H}}
\newcommand{\groupclassifier}[1]{h^g_{#1}}
\newcommand{\kldiv}{D_\text{KL}}

\newcommand{\salun}{\textsc{SalUn} \citep{fan2023salun}}
\newcommand{\sparse}{\textsc{L1-sparse} \citep{jia2024modelsparsitysimplifymachine}}
\newcommand{\scrub}{\textsc{SCRUB} \citep{kurmanji2024towards}}
\newcommand{\dro}{\textsc{group-DRO} \citep{sagawa2019distributionally}}

\newcommand{\reweight}{\textsc{reweight}}
\newcommand{\pretrain}{\textsc{Pretrain}}
\newcommand{\retrain}{\textsc{Retrain}}
\newcommand{\method}{\textsc{MIU}} %
\newcommand{\fullname}{\textsc{\underline{M}utual \underline{I}nformation-aware Machine \underline{U}nlearning}}

\newcommand{\celeba}{CelebA \citep{liu2015faceattributes}}
\newcommand{\fairface}{FairFace \citep{karkkainenfairface}}
\newcommand{\waterbird}{Waterbirds \citep{sagawa2019distributionally}}
\newcommand{\multinli}{MultiNLI \citep{N18-1101}}

\newcommand{\gap}{avg.\ gap}

\definecolor{colorexp}{RGB}{237, 242, 244}
\definecolor{colormethod}{HTML}{c7dcfc}
\definecolor{colorgap}{HTML}{dddddd}
\definecolor{colorfunc}{HTML}{8338ec}
\definecolor{colorcomment}{HTML}{3a86ff}
\newcommand{\targetattr}[1]{\textcolor{OrangeRed}{#1}}
\newcommand{\sensitiveattr}[1]{\textcolor{RoyalBlue}{#1}}

\newcommand{\rowmethod}{\rowcolor{colormethod}}
\newcommand{\inlinecolorbox}[2]{\begingroup\setlength{\fboxsep}{1pt}\colorbox{#1}{\hspace{2pt}\vphantom{Ay}#2\hspace{2pt}}\endgroup}

\newcommand{\func}[1]{\textcolor{colorfunc}{#1}}
\newcommand{\comm}[1]{\textcolor{colorcomment}{\# #1}}

\def\checkmark{\tikz\fill[scale=0.3](0,.35) -- (.25,0) -- (1,.7) -- (.25,.15) -- cycle;}

\crefname{section}{Sec.}{Secs.}
\crefname{table}{Tab.}{Tabs.}
\crefname{figure}{Fig.}{Figs.}
\crefname{appendix}{Appx.}{Appxs.}
\crefname{equation}{Eq.}{Eqs.}
\Crefname{section}{Section}{Sections}
\Crefname{table}{Table}{Tables}
\Crefname{figure}{Figure}{Figures}
\Crefname{appendix}{Appendix}{Appendices}
\Crefname{equation}{Equation}{Equations}

\title{Group-robust Machine Unlearning}

\author{\name Thomas De Min \\
    \addr University of Trento, Italy \\
    \email thomas.demin@unitn.it
    \AND
    \name Subhankar Roy \\
    \addr University of Bergamo, Italy
    \AND
    \name Stéphane Lathuilière \\
    \addr Inria Grenoble, Univ.\ Grenoble Alpes, France
    \AND
    \name Elisa Ricci\\
    \addr University of Trento, Italy \\
    \addr Fondazione Bruno Kessler, Italy
    \AND
    \name Massimiliano Mancini \\
    \addr University of Trento, Italy
}

\def\month{12}  %
\def\year{2025} %
\def\openreview{\url{https://openreview.net/forum?id=StSq7mpUVw}} %

\begin{document}

\maketitle

\begin{abstract}
Machine unlearning is an emerging paradigm to remove the influence of specific training data (\ie, the forget set) from a model while preserving its knowledge of the rest of the data (\ie, the remaining set).
Previous approaches assume the forget data to be uniformly distributed from all training datapoints.
However, if the data to unlearn is dominant in one \textit{group} (\eg, ethnicity, gender), we empirically show that performance for this group \changed{can degrade}, leading to fairness issues.
To perform unlearning while preserving fairness, this work addresses the overlooked problem of non-uniformly distributed forget sets, which we refer to as \textit{group-robust machine unlearning}. 
We formalize the problem and present a simple and effective exact unlearning strategy that mitigates the performance loss in dominant groups via sample distribution reweighting.
Moreover, we present \method{} (\fullname{}), the first approach for group robustness in approximate machine unlearning. 
\method{} minimizes the mutual information between model features and group information, achieving unlearning while reducing performance degradation in the dominant group of the forget set. 
Additionally, \method{} exploits sample distribution reweighting and mutual information calibration with the original model to preserve group robustness.
We conduct experiments on three datasets and show that \method{} outperforms standard methods, achieving unlearning without compromising model robustness.
Source code available at \href{https://github.com/tdemin16/group-robust_machine_unlearning}{\small{\url{https://github.com/tdemin16/group-robust_machine_unlearning}}}.
\end{abstract}

\section{Introduction}
\label{sec:intro}
In several countries, recent regulations grant users greater control over their digital privacy, explicitly recognizing their right to request the removal of personal data, known as the ``right to be forgotten'' \citep{mantelero2013eu,voigt2017eu}.
Therefore, to comply with such regulations, machine learning systems should be able to forget specific training data upon user request—a challenge addressed by machine unlearning \citep{kurmanji2024towards,chundawat2023can}.

In the machine unlearning literature, the standard setup assumes that a group of individuals requests the removal of their data. A machine unlearning algorithm then fulfills this request by making the model ``forget'' the designated data (\ie, the forget set) while maintaining its utility \citep{kurmanji2024towards}.
The naive solution, called \emph{exact unlearning}, consists of retraining the model without the forget data; however, this is computationally prohibitive for large models \citep{zhao2024makes}.
\emph{Approximate unlearning} methods \citep{jia2024modelsparsitysimplifymachine,fan2023salun,kurmanji2024towards} overcome this limitation by unlearning the model with significantly fewer resources but with less unlearning guarantees.

\begin{figure}
    \centering
    \includegraphics[width=\linewidth]{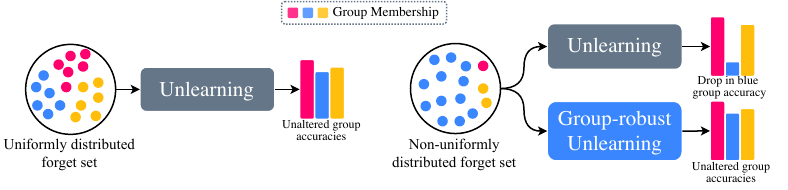}
    \caption{\textbf{Comparing unlearning approaches.} Previous works assume the forget set to be uniformly distributed. However, real-life unlearning requests do not comply with the uniform distribution assumption \citep{bertram2019five}. If the forget set distribution is predominant in some groups (\eg, old males), it can lead to performance degradation in such dominant forget groups (\ie, the blue group in the figure).
    Group-robust Unlearning prevents this from happening.
    }
    \label{fig:teaser}
\end{figure}

Most approaches assume that the forget sets are uniformly sampled from the training data \citep{jia2024modelsparsitysimplifymachine,fan2023salun,cheng2023multimodal,shen2024label,he2024towards,zhao2024makes,huang2024learning}. 
However,  studies \citep{bertram2019five,zhang2024forgotten} show that individuals from different social and cultural backgrounds (\ie, \emph{groups}) request to be forgotten at varying rates: \citet{bertram2019five} reports that 44.4\% of unlearning requests to news websites relate to professional wrongdoing or crimes, while \citet{zhang2024forgotten} finds that wealthier, highly educated individuals are more likely to request unlearning.  
Existing methods overlook this imbalance, potentially degrading the accuracy of dominant groups in the forget set and leading to unfair outcomes (see~\cref{fig:teaser,fig:ratio}).
This can be critical in scenarios requiring high accuracy across all groups.
In a recommendation system, if unlearning requests predominantly come from a specific group (\eg, young males), the system's quality for that group may degrade, potentially making it unusable. Thus, machine unlearning algorithms must account for distribution shifts.

We target this unexplored scenario, which we name \textit{group-robust machine unlearning}, that aims to unlearn the user's data while minimizing the performance deterioration of groups dominating the forget set.
Differently from previous works, we tackle both \emph{exact} and \emph{approximate} unlearning by (i) finding a retraining strategy that preserves the original group robustness \citep{sagawa2019distributionally}, and (ii) 
proposing an approximate unlearning method that efficiently forgets data while preserving robustness for unbalanced forget sets.

For (i), we show that reweighting the sampling distribution during retraining compensates for information loss with minimal robustness impact.
We validate this strategy (called \reweight{}) against \dro{}, a popular group-robust optimization method, showing that \reweight{} better preserves model group robustness in exact unlearning.
For (ii), we introduce an approximate unlearning method called \method{} (\fullname{}), leveraging mutual information minimization \citep{belghazi2018mine} and calibration to unlearn the forget set while preserving model robustness. 
By minimizing the mutual information between forget-set features and ground-truth group annotations, we decorrelate unlearning from spurious attributes \citep{liu2021just}, mitigating performance loss for dominant groups.
To prevent affecting other groups, we calibrate the unlearned model’s mutual information to match the original one.
Coupled with \reweight{}, \method{} outperforms established unlearning approaches (\sparse{}, \salun{}, and \scrub{}) on \celeba{}, \waterbird{}, and \fairface{} in both unlearning efficacy and preserved group robustness.\footnote{We use robustness and group-robustness interchangeably in this paper.}

\noindent\textbf{Contributions.} In summary, our contributions are:
\begin{itemize}
    \item We are the first to identify the issue of group robustness in approximate unlearning, showing how existing unlearning algorithms degrade model robustness in this setting.
    \item We propose a simple and effective sample distribution reweighting strategy to mitigate the group accuracy degradation in exact unlearning.
    \item We introduce \method{}, the first approximate unlearning approach tailored for this task that minimizes the mutual information on the forget set while calibrating it to match the group robustness of the original model.
    \item We benchmark existing baselines and \method{} on group-robust machine unlearning using \celeba{}, \waterbird{}, and \fairface{}, and showing that \method{} outperforms existing methods in this task.
\end{itemize}

\section{Related work}
\noindent\textbf{Machine unlearning.}
Exact machine unlearning methods \citep{bourtoule2021machine,yan2022arcane,aldaghri2021coded} guarantee that sensitive data is removed.
However, they are impractical \citep{chundawat2023can,kurmanji2024towards} as retraining (part of) the model is prohibitive \citep{nguyen2022survey}.
Instead, approximate unlearning concentrates on computational feasibility by relaxing the guarantees constraints \citep{kurmanji2024towards,chundawat2023can,jia2024modelsparsitysimplifymachine,fan2023salun,chen2023boundary}.
Most works focus on random \citep{jia2024modelsparsitysimplifymachine,fan2023salun,cheng2023multimodal,shen2024label,he2024towards,zhao2024makes,huang2024learning}, and class \citep{chen2023boundary,chundawat2023can,chundawat2023zero,hoang2024learn,cheng2024machine,zhao2024continual} unlearning, respectively forgetting i.i.d.\ data points, and all data points from a single class.
While the former lowers the forget-set accuracy to match that of the model retrained without the unlearning data, the latter aims at scoring zero accuracy on the unlearned class.
A few works \citep{chundawat2023can,foster2024fast,cheng2024machine} explore subclass unlearning, where a subset of a class (\eg, subclass \textit{car} from superclass \textit{vehicles}) is removed from model weights.
While subclass and group-robust unlearning share similarities, the latter focuses on preserving group accuracies, whereas subclass unlearning alters the model to behave as if the target subclass was never in the training set.
Related to our research, \citet{fan2024challenging} suggests a bi-level optimization to sample adversarial forget sets that are difficult to unlearn.
Also related, \citet{chen2024fast} proposes machine unlearning as an efficient debiasing technique.
Instead, \citet{zhang2024forgotten} investigates how uniform and non-uniform data sampling affect fairness in MLPs and tabular data, limiting the evaluation to exact unlearning.

\changed{Group-robust unlearning can be seen as a generalized case of random unlearning, where we drop the i.i.d.\ assumption of the forget set.
This has implications for the design of proposed methodologies.
As existing approaches assume a uniform distribution of the forget set during evaluation, they do not account for group imbalance in the forget distribution, possibly degrading the model's robustness.}

\noindent\textbf{Group-robust learning.}
Methods for group-robust optimization train deep learning models to be robust to spurious correlations.
Algorithms are categorized based on their access to group information.
Within those that assume access to group annotations \citep{sagawa2019distributionally,goel2020model,zhang2020coping,idrissi2022simple,kirichenko2022last},
group-DRO \citep{sagawa2019distributionally} dynamically reweights the misclassification penalty for each group to optimize the worst-group accuracy.
Instead, \citet{idrissi2022simple} proposes a simple baseline that subsamples each group to match the size of the smallest one.
While these works usually achieve better results, accessing the group information can be challenging (\eg, annotation cost).
Within methods agnostic to group information \citep{liu2021just,zhang2022correct,idrissi2022simple,sohoni2020no,nam2020learning}, JTT \citep{liu2021just} increases the sampling probability of wrongly classified data to improve worst-group accuracy.
Correct-N-Contrast \citep{zhang2022correct} uses a contrastive loss to pull correctly and misclassified samples of the same class while pushing apart wrongly classified data points of different categories.
However, methods that require group information also show state-of-the-art performance on groups discovered from data \citep{kim2024discovering,d2024openbias}.

This paper is the first to study the intersection between group robustness and machine unlearning. 
For this reason, we assume complete access to the group information.

\section{Method}
\label{sec:task}
This section formulates the machine unlearning task~(\cref{sub:problem}) and the group-robust machine unlearning problem~(\cref{sub:harm}).
Then, it introduces the proposed sample distribution reweighting strategy~(\cref{sub:naive}), showing its effectiveness in group-robust machine unlearning.
Finally, \cref{sub:method} describes \method{}, our approximate unlearning method tailored for group-robust unlearning.

\subsection{Machine unlearning}
\label{sub:problem}
Let~$\model:X\rightarrow\targets$ be a learnable function (or model), where~$f_\theta(\cdot):X\rightarrow Z$ is a non-linear feature extractor parameterized by $\theta$, and~$h_\varphi(\cdot): Z\rightarrow\targets$ is a linear classifier parameterized by $\varphi$, mapping inputs from the image $X$ to the target space $\targets$.
Let $\trainds = \{(\image, \target)_i\}_{i=1}^{N_{\mathit{tr}}}$ be a training dataset of size~$N_{\mathit{tr}}$, where~$\image_i$ is an image, {and }$\target_i$~its target label (\eg, age).
\changed{Let a model trained on $\trainds$ with algorithm $\training$ be denoted as $\original$ (or \pretrain{}). A machine unlearning algorithm $\mathcal{U}$ scrubs the influence of a desired forget set $\forgetds\subset\trainds$ from the pretrained model by outputting scrubbed weights $\{\varphi_u$, $\theta_u\}$, such that $\unlearned$ is as close as possible to the exact unlearning model $\retrained$ (or \retrain{}), trained solely on the remaining set $\retainds = \trainds\setminus\forgetds$ with algorithm $\training$ \citep{kurmanji2024towards,fan2023salun,zhao2024makes}.}

\subsection{Group-robust machine unlearning}
\label{sub:harm}
\begin{wrapfigure}[19]{r}{0.5\linewidth}
    \vspace{-\baselineskip}
    \centering
    \includegraphics[width=\linewidth]{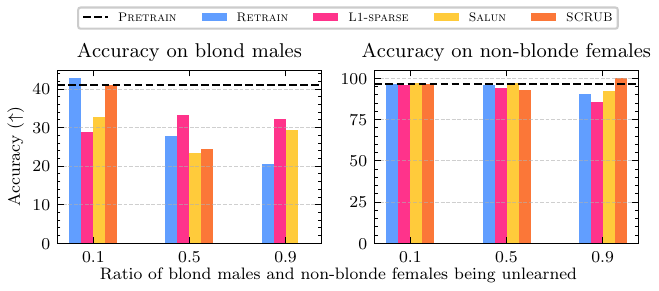}
    \caption{
        \changed{\textbf{Unlearning non-uniformly distributed data.}
        We test standard model retraining, and popular approximate unlearning methods (\textsc{L1-sparse}, \textsc{SalUn}, and \textsc{SCRUB}) in group-robust unlearning.
        The more samples of the least represented group (\emph{\targetattr{blond} \sensitiveattr{males}}) are unlearned from CelebA, the lower the model accuracy on that group. On the contrary, the most represented one (\emph{\targetattr{non-blonde} \sensitiveattr{females}}) is less affected.}
    }
    \label{fig:ratio}
\end{wrapfigure}
In group-robust machine unlearning, we consider the forget data non-uniformly distributed.
Therefore, let us re-define the training dataset as $\trainds = \left\{(\image, \target, \attribute)_i\right\}_{i=1}^{N_{\mathit{tr}}}$, where $\image_i$ and $\target_i$ are defined as in \cref{sub:problem}, and $\attribute_i$ is the protected or sensitive attribute (\eg, \textit{gender}, \textit{ethnicity}).
Now, let $\groups:\targets\times\attributes$ be the set of all groups, defined as the cartesian product between the target label set $\targets$ and the protected attribute set $\attributes$ \citep{sagawa2019distributionally,kirichenko2022last}.
We denote the $i$\textit{-th} datapoint group as $g_i=(\targetattr{\target_i}, \sensitiveattr{\attribute_i})$.
Each target-sensitive attribute pair (\eg, \sensitiveattr{\textit{males}}, between the ages of \targetattr{\textit{20-29}}) identifies a unique group.\footnote{\targetattr{Target} and \sensitiveattr{protected} attributes are color-coded.}

\changed{If the forget set is non-i.i.d., the group distribution in the remaining set changes compared to the original training set.
As the proportion of a group in the remaining set lowers, its resulting accuracy also decreases, which can harm} the model's generalization performance on the dominant group of the forget set.
\changed{\Cref{fig:ratio} shows the accuracy degradation caused by jointly removing different percentages of the least and most represented groups, \ie, \emph{\targetattr{blond} \sensitiveattr{males}} and \emph{\targetattr{non-blonde} \sensitiveattr{females}}.
While the most represented group is mildly affected as its ratio drops from 44\% to 7\% (over four groups), the least represented one (dropping from 0.8\% to 0.1\%) suffers a substantial accuracy degradation of $\sim$20\%.}
Further analysis is provided in \cref{appx:additional_results}.

\changed{Therefore, the objective of group-robust machine unlearning differs from that of machine unlearning. While the latter minimizes the discrepancy with the retrained model (gold standard), the former also preserves the original model's performance on each group.}

\subsection{Frustratingly easy group-robust unlearning}
\label{sub:naive}
\Cref{sub:harm} introduces the group-robust machine unlearning task and shows the extent to which the accuracy of the forget set dominant group(s) drops after unlearning.
Unlike prior works (see \cref{sec:intro}), we aim at unlearning non-uniform forget data while preserving the model accuracy on the dominant group of the forget set.
Therefore, this section proposes an exact unlearning strategy tailored to this task.
To mitigate the performance degradation of the dominant group of $\forgetds$, we argue that reweighting the data distribution \changed{(RWG in \citet{idrissi2022simple})} to account for the removed samples is a simple and effective baseline that retrains the model with a minimal performance drop.
\changed{Contrary to RWG \citep{idrissi2022simple}, which enforces uniform sampling distribution from group-perspective, we suggest instead to reweight the sampling distribution to match the original dataset statistics.}
Intuitively, increasing the sampling likelihood for partly unlearned groups rebalances the remaining set group statistics to match those of the training dataset, recovering the original robustness.

Formally, let $\p(\image_i) = \frac{1}{N_{\mathit{tr}}}$ be the probability of sampling~$\image_i$, let $\nu_{\mathit{tr}}, \nu_{\mathit{r}} \in \mathbb{N}^{\groups}$ respectively be the group frequencies of the training and remaining datasets.
We reweight $\p(\image_i)$ according to the ratio $\alpha = \frac{\nu_{\mathit{tr}}}{\nu_{\mathit{r}}}$: $\p(\image_i) = \frac{\alpha_{g_i}}{\sum_j\alpha_{g_j}}$, where~$\alpha_{g_i}$ is the adjusted group weight for sample~$\image_i$.
We denote the data distribution reweighting strategy (or \reweight{}) as~$\omega(\cdot)$, \ie, $\omega(\retainds)$ denotes the reweighting strategy applied to the remaining dataset.

\begin{wrapfigure}[16]{r}{0.5\linewidth}
    \vspace{-\baselineskip}
    \centering
    \includegraphics[width=\linewidth]{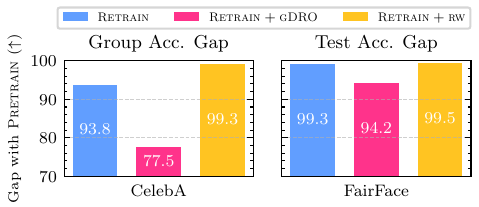}
    \caption{\textbf{\reweight{} \vs{} \textsc{group-DRO}.} \retrain{} + \reweight{} achieves a better test and group accuracy alignment with the original model (higher is better). Thus, it better preserves the original performance after unlearning.}
    \label{fig:rob}
\end{wrapfigure}

We validate \reweight{} by comparing the model retrained with \reweight{} and \dro{}.
\Cref{fig:rob} summarizes the outcome of our analysis.
As a byproduct of strongly optimizing worst-group accuracies, we notice that \dro{} can also increase the forget set accuracy.
This issue makes approximate unlearning evaluation more difficult if \retrain{} + \textsc{group-DRO} is used as the gold standard.
Assuming a hypothetical original forget-set accuracy of 70\%, if an approximate unlearning algorithm targeting such a gold standard leads to higher accuracy (\eg, 80\%), then \emph{was the knowledge unlearned if forget-set accuracy increased?}
Answering this question is non-trivial, as the retrained model cannot be accessed for real applications.
Furthermore, retraining with \dro{} causes a test accuracy drop (-5.8\%) in \fairface{}.
On the contrary, \reweight{} does not suffer from these issues.

\subsection{Group-robust unlearning without retraining}
\label{sub:method}
\Cref{sub:naive} shows that sample reweighting is a simple and effective approach to machine unlearning that preserves the original model's robustness.
However, model retraining is inefficient \citep{chundawat2023can,kurmanji2024towards} and unpractical \citep{nguyen2022survey}.
Therefore, this section proposes \method{}, our approximate machine unlearning algorithm that jointly tackles unlearning while decorrelating unlearning and group robustness without retraining the entire model.

As \textit{scrubbing} unlearning data affects the forget set dominant group accuracy, we propose a unified objective for jointly unlearning while preserving original group robustness.
We use mutual information between output features and group annotation, which, upon minimization on forget data, jointly unlearns and mitigates the performance degradation on the dominant group of the forget set \citep{ragonesi2021learning}.
Formally, let $I(Z; \groups) = I(Z; (\targets, \attributes))$ be the mutual information between random variables $Z$ and $\groups$, associated with model features $\mathbf{z} = \backbone{\theta}(\image)$ and group $g=(\target,\attribute)$, then:
\begin{equation}
    \label{eq:mi}
    I(Z; \groups) = \int_\groups\int_Z \mathbb{P}_{(Z, \groups)}(\mathbf{z}, g)\log\left( \frac{\mathbb{P}_{(Z, \groups)}(\mathbf{z}, g)}{\mathbb{P}_Z(\mathbf{z}) \mathbb{P}_\groups(g)} \right) d\mathbf{z}~dg\text{,}
\end{equation}
where $\mathbb{P}_{(Z, \groups)}$ is the joint pdf of $Z$ and $\groups$, and $\mathbb{P}_Z$, $\mathbb{P}_\groups$ are marginal pdfs of $Z$ and $\groups$.
Minimizing $I(Z, \groups)$ reduces the dependency between $Z$ and $\targets$ while also reducing it between $Z$ and $\attributes$, jointly unlearning the network and disentangling features and protected attributes.
However, computing the mutual information for continuous variables is generally intractable \citep{paninski2003estimation,belghazi2018mine}.
Therefore, we follow \citet{belghazi2018mine}, and estimate the mutual information with an MLP $T_\psi$:
\begin{equation}
\label{eq:mine}
    \mine(\dataset; \psi, \theta) = \mathbb{E}_{(\image, g)\sim\dataset}T_\psi(\mathbf{z}, g) - \log\mathbb{E}_{(\image, \bar{g})\sim\dataset}\left[e^{T_\psi(\mathbf{z},\bar{g})}\right]\text{,}
\end{equation}
where \changed{$T_\psi$ is a two-layer MLP with ReLU as non-linear activation, outputting a single scalar}, $(\image, g)\sim\dataset$ refers to a sampling from the joint distribution, $(\image, \bar{g})\sim\dataset$ refers to a sampling from the product of the marginal distributions.
To sample from the product of the marginal distributions we sample twice from the joint distribution: $(\image,\bar{g})\,\text{s.t.}\,(\image,g)\sim\dataset,(\bar{\image},\bar{g})\sim\dataset$, and keep only $\image$ and $\bar{g}$ as in \citet{belghazi2018mine}.
We now use this definition to derive \method{} (\fullname{}), our proposed method.

\noindent\textbf{Unlearning term.}
We minimize the mutual information between forget set features and their group label to unlearn the forget set $\forgetds$ while maintaining a good trade-off between information removal and robustness preservation.
Therefore, we denote the unlearning term as:
\begin{equation}
\label{eq:unlearn}
    \unlearningterm.
\end{equation}
This term is high when the forget features correlate with the group information. 
Intuitively, we want this term to be low as this implies that we are unlearning the relation between forget-set features and labels while decorrelating the unlearning process from group information.
Formally, minimizing this term achieves unlearning while preserving group robustness due to the following proposition:
\begin{proposition}
    Mutual information minimization is equivalent to cross-entropy loss maximization on a group classifier.
\end{proposition}
\begin{proof}
Let us rewrite \cref{eq:mi} in terms of entropy and conditional entropy:
\begin{equation}
    \label{eq:mi_entropy}
    I(Z;\groups) = \entropy(\groups) - \entropy(\groups | Z) \propto -\entropy(\groups | Z)\text{,}
\end{equation}
where group entropy $\entropy(\groups)$ is constant, thus, it can be ignored during optimization.
Let $\groupclassifier{\xi}(\mathbf{z}) = \hat{g}$ be a group classifier, namely a linear layer that classifies image groups using features $\mathbf{z}$, and let $\widehat{\groups}$ be the random variable associated with estimated groups.
We can now write the cross-entropy loss for classifying groups using entropy notation as:
\begin{equation}
    \label{eq:group_ce}
    \entropy(\groups;\widehat{\groups}|Z) = \entropy(\groups|Z) + \kldiv(\groups \| \widehat{\groups} | Z)\text{.}
\end{equation}
Following \citet{boudiaf2020unifying}, Lemma 2, we can relate \cref{eq:group_ce,eq:mi_entropy} by decoupling \cref{eq:group_ce} into two steps, \ie, first optimize $\kldiv(\groups \| \widehat{\groups} | Z)$ and then $\entropy(\groups|Z)$.
In other words, the first step fixes the backbone weights $\theta$ and optimizes the group classifier weights $\xi$, while the second only optimizes model weights $\theta$, keeping the classifier weights frozen.
As the classifier is fixed in the latter, then maximizing the cross-entropy with respect to encoder weights corresponds to maximizing $\entropy(\groups|Z)$, \ie, minimizing $I(Z;\groups)$.
Thus, minimizing $I(Z;\groups)$ (as in \cref{eq:unlearn}) is equivalent to computing gradient ascent on group information, functionally disentangling label and spurious attributes from image features.
\end{proof}

\noindent\textbf{Calibration term.}
As unlearning might also affect other groups, we designed an extra term to improve group performance retention.
Thus, we minimize the mutual information discrepancy between the original and unlearning model on the reweighted remaining dataset $\reweighting(\retainds)$:
\begin{equation}
\label{eq:regularize}
    \alignmentterm = \left\| \mine(\reweighting(\retainds); \psi, \theta)\, - \,\mine(\reweighting(\retainds); \psi, \theta_o)\right\|^2_\text{,}
\end{equation}
where $\mine(\reweighting(\retainds); \psi, \theta_o)$ estimates the mutual information using features computed by the original backbone $\backbone{\theta_o}(\cdot)$. 
\Cref{eq:regularize} encourages the unlearned model to mimic the original robustness, preserving its behavior across groups.

\noindent\textbf{Retaining term.}
Since unlearning generally causes performance degradation \citep{kurmanji2024towards,fan2023salun,jia2024modelsparsitysimplifymachine}, we perform gradient descent steps on the remaining data to ensure that the original model group accuracy is preserved after unlearning.
Additionally, we use \reweight{} to limit the degradation of group robustness.
Therefore, we optimize the cross-entropy loss on the reweighted remaining dataset $\reweighting(\retainds)$:
\begin{equation}
\label{eq:retain}
    \retainingterm = {\mathbb{E}_{\sampling}\,\mathcal{L}_\text{CE}(\image, \target; \varphi, \theta)}\text{.}
\end{equation}

\noindent\textbf{Putting all together.}
The \emph{retaining term} in \method{} preserves the model's original discriminative capabilities.
The \emph{unlearning term} removes $\forgetds$ from model weights while minimizing performance loss for dominant groups in the forget set.
Finally, the \emph{calibration term} ensures that the unlearned model maintains its original robustness.
Therefore:
\begin{equation}
\label{eq:full}
    \varphi_u, \theta_u = \optim\,\underset{\text{Retaining term}}{\underbrace{\retainingterm}}\, +\, \underset{\text{Unlearning term}}{\underbrace{\unlearningterm}}\, +\, \lambda\cdot\underset{\text{Calibration term}}{\underbrace{\alignmentterm}} \text{.}
\end{equation}

\noindent\textbf{\method{} pseudocode.}
\Cref{algo:pseudocode} presents \method{} pseudocode in a PyTorch-like \citep{paszke2019pytorch} style.
We follow previous works \citep{kurmanji2024towards,fan2023salun} and compute \textit{unlearning} and \textit{retaining} steps separately.
Thus, we alternate between computing an unlearning epoch using \cref{eq:unlearn} and a retaining one using \cref{eq:retain,eq:regularize}.
Like \scrub{}, we find it beneficial to stop performing unlearning steps after a predefined number of epochs (\ie, 5 out of 10).
Contrary to \citet{ragonesi2021learning}, we do not update the mutual information representation at every step.
Instead, we empirically observed that updating it via 100 gradient updates for the first epoch and 10 updates for the remaining iterations is sufficient to achieve satisfactory results and limit the required resources.
As we keep the optimization steps fixed, the overall mutual information estimation overhead depends on the dataset size.
For a small dataset like \waterbird{}, we estimate an overhead of about $1.80\times$.
Instead, for \celeba{} and \fairface{}, we estimate an increase of approximately $1.03\times$ in unlearning time, which is negligible.
\begin{algorithm*}[htp]
    \caption{PyTorch-like \method{} pseudocode.}
    \begin{algorithmic}
        \tt
        \footnotesize
        \State def \func{MIU}(model, mine, mine\_original, optim, train\_dataloader, remaining\_dataloader, forget\_dataloader):
        \State \qquad unlearned = copy.\func{deepcopy}(model)
        \State \qquad for epoch in \func{range}(epochs):
        \State \qquad \qquad \comm{Update mine (mutual information neural estimator) as in} \citet{belghazi2018mine}
        \State \qquad \qquad \func{tune\_mine}(unlearned, mine, train\_dataloader)
        \State
        \State \qquad \qquad \comm{Use mine to unlearn the forget\_dataset for the first forget\_epochs epochs}
        \State \qquad \qquad if epoch < forget\_epochs:
        \State \qquad \qquad \qquad for image, group in forget\_dataloader:
        \State \qquad \qquad \qquad \qquad group\_bar = torch.\func{randint\_like}(group, num\_groups)
        \State \qquad \qquad \qquad \qquad loss = \func{mine}(\func{unlearned}(image), group, group\_bar)
        \State \qquad \qquad \qquad \qquad \func{update\_model}(loss, optim)
        \State
        \State \qquad \qquad \comm{Apply fine-tuning steps at every epoch using the reweighted sampler}
        \State \qquad \qquad for image, target, group in \func{reweight}(remaining\_dataloader):
        \State \qquad \qquad \qquad group\_bar = torch.\func{randint\_like}(group, num\_groups)
        \State \qquad \qquad \qquad loss = F.\func{cross\_entropy}(\func{head}(\func{unlearned}(image)), target)
        \State \qquad \qquad \qquad loss += lambda * F.\func{mse\_loss}(
        \State \qquad \qquad \qquad \qquad \func{mine}(\func{unlearned}(image), group, group\_bar),
        \State \qquad \qquad \qquad \qquad \func{mine\_original}(\func{model}(image), group, group\_bar).\func{detach}()
        \State \qquad \qquad \qquad )
        \State \qquad \qquad \qquad \func{update\_model}(loss, optim)
        \State \qquad return unlearned
    \end{algorithmic}
    \label{algo:pseudocode}
\end{algorithm*}

\section{Experiments}
\label{sec:experiments}
This section describes the experimental protocol~(\cref{sub:procedure}) and compares \method{} with multiple methods in group robust unlearning~(\cref{sub:results}).
We then compare existing approaches and ours when varying the unlearning ratio of the forget set dominant group (\cref{sub:varying}) and when sampling the forget set from multiple groups (\cref{sub:multigroup}).
Finally, \cref{sub:ablations} shows a complete ablation study of \method's components.

\subsection{Experimental protocol}
\label{sub:procedure}
\textbf{Datasets.}
We follow established works in group-robust optimization \citep{sagawa2019distributionally,idrissi2022simple,liu2021just,park2022fair,park2024fair} and conduct experiments on \celeba{} and \waterbird{} datasets, setting \targetattr{\textit{blonde}} and \sensitiveattr{\textit{male}} as the target and protected attributes for \celeba{}, while setting \targetattr{\textit{waterbirds}} and \textit{\sensitiveattr{land}} as the target and sensitive attributes for \waterbird{}.
We additionally experiment with \fairface{}, given its numerous annotations, using age as the downstream task and randomly choosing the class \targetattr{\textit{20-29}} and the ethnicity \sensitiveattr{\textit{afro-american}} as target and protected attributes, respectively.
Unless stated otherwise, forget-set images are sampled from groups described by the above target-sensitive attribute pairs.
Moreover, unless stated otherwise, the forget set simulates the worst-case scenario where a single group is responsible for the unlearning request, leading to a high forget distribution imbalance.
\changed{The unlearning ratio is defined as the proportion of samples from that particular group that have been unlearned.}
After unlearning, the model must have unlearned the forget data and maintained its original robustness.

\noindent\textbf{Hardware and training details.}
We obtain \pretrain{} and \retrain{} by fine-tuning \changed{via empirical-risk minimization} a ResNet-18 \citep{he2016deep} pre-trained on ImageNet \citep{russakovsky2015imagenet} for 30 epochs, using SGD with 0.9 momentum and weight decay.
The learning rate is decayed with a cosine annealing scheduler for the entire training.
We additionally warm-up the learning rate for the first two epochs using a linear scheduler.
We apply standard data augmentation techniques, namely, random resized crop, random horizontal flip, and input normalization \citep{he2016deep}.
We limited fine-tuning to 10 epochs for approximate unlearning methods, searching for the optimal configuration for the other hyperparameters.
The $\lambda$ parameter of \method{} is set between 1 and 10 (see~\cref{appxsub:ablations}).
All experiments ran on a single A100 Nvidia GPU, using PyTorch \citep{paszke2019pytorch}. 

\noindent\textbf{Baselines.}
We compare \method{} against three state-of-the-art machine unlearning approaches.
\sparse{} forgets sensitive information by fine-tuning the original model on the remaining set with a sparsity regularization term.
\salun{} proposes a saliency-based unlearning that forgets data via random labeling.
\scrub{} minimizes the divergence between the unlearned and original model on the remaining set while maximizing it on the unlearning data.
Following previous works \citep{kurmanji2024towards,jia2024modelsparsitysimplifymachine}, we report \pretrain{}, and \retrain{}, which are computed by fine-tuning an ImageNet \citep{russakovsky2015imagenet} pre-trained ResNet-18 \citep{he2016deep}.
We also report \retrain{} + \dro{} to validate the proposed \retrain{} + \reweight{}. 

We reimplemented all three %
baselines following the existing codebases.
For \sparse{}, we perform 10 fine-tuning epochs on the remaining set with a linearly decaying $L1$ regularization that follows this rule: $\gamma_t = (1 - t/T)\gamma$, where $t$ is the epoch, $T$ is the total number of iterations, and $\gamma$ the initial penalty.
For \salun{} we followed the ``small-scale'' implementation of the original codebase (\ie, the implementation for CIFAR-10 \citep{krizhevsky2009learning} and SVHN \citep{svhn}) as it is meant for datasets with few classes.
We first compute the saliency mask using the gradient information from the forget set, pruning 50\% of the network weights.
We tune the pruned weights for 10 epochs by alternating a full pass on the forget set and a full pass on the remaining set.
Also, \scrub{} is implemented by separating forget and retaining steps, following the original code.
The forget step maximizes the KL divergence between the original and the unlearned model in the forget data set. 
Following the original paper, we stop computing it after 3, 5, or 7 epochs.
Instead, the retaining step minimizes the linear combination between the cross-entropy loss and the KL divergence between the original and unlearned model, respectively scaled by 0.99 and 0.001, as reported in \citep{kurmanji2024towards}.
We note that all methods use the same dataset splits; therefore, they must unlearn the same forget set.

\noindent\textbf{Metrics.}
To evaluate unlearning and group-robustness, we rely on six different metrics. 
The first three are remaining (RA), forget (UA), and test (TA) accuracy.
We also report the membership inference attack (MIA) \citep{yeom2018privacy}, which measures the MIA-Efficacy as described in \citep{jia2024modelsparsitysimplifymachine,fan2023salun}.
Finally, we assess the change in group robustness by looking at equalized odds (EO) \citep{hardt2016equality}, and the test accuracy of the forget-set dominant group (GA).

\changed{GA measures the ratio of correctly classified test samples that belong to the same dominant group(s) of the forget set.
Therefore, let $\testds^{g_\mathit{f}} = \{(\image_i, \target_i, \attribute_i) \mid g_\mathit{f} = g_i = (\target_i, \attribute_i) \wedge (\image_i, \target_i, \attribute_i) \in \testds\}$ be the subset of the test set composed of all samples of group $g_\mathit{f}$, then GA is computed as follows:}
\begin{equation}
    \text{GA} = \frac{1}{|\testds^{g_\mathit{f}}|} \sum_{i=1}^{|\testds^{g_\mathit{f}}|} \mathds{1}\left[ (\model)(\image_i) = \target_i \right]\text{,}
\end{equation}
\changed{where $\mathds{1}$ is the indicator function, returning $1$ if the argument is True, and $0$ otherwise.
Like other accuracy metrics, the closer GA is to 1 (or 100\%), the better the model is at classifying data of the dominant group of the forget set.}

Finally, to ease the interpretation of six metrics, we follow previous works \citep{jia2024modelsparsitysimplifymachine,fan2023salun} and compute the average gap (\textbf{\gap{}}) and per-metric deltas with the gold standard.\footnote{Deltas are computed per each seed and then averaged over three runs.}
Following them, we do not report whether metrics should be maximized or minimized as the machine unlearning objective is to reduce the discrepancy with the gold standard on each metric, except for \gap{}, which must be maximized.
We use \retrain{} + \reweight{} as the gold standard since it better reduces the gap with \pretrain{} in TA, EO, and GA, compared to \retrain{} and \retrain{} + \dro{}.
Further details are in \cref{appx:metrics}.

\subsection{Results on group unlearning}
\label{sub:results}
\begin{table*}[t]
    \scriptsize
    \centering
    \caption{\changed{\textbf{Group-robust machine unlearning in \celeba{}.} We build the forget set by sampling data points from a single group. The unlearning ratio is set to 0.5. We compare \inlinecolorbox{colormethod}{\method{}} against \sparse{}, \salun{}, and \scrub{}. The \gap{} and deltas are computed against \retrain{} + \reweight{}, and we bold the methods that achieve the smallest discrepancy with it. Other reference models are in \textcolor{gray}{light gray}.}}
    \vspace{\baselineskip}
    \begin{tabularx}{\textwidth}{
        L{4cm}
        C{0.5cm}
        *{6}{Y}
        C{1.5cm}
    }
        method & RW & RA & UA & TA & MIA & EO & GA & \textbf{\gap{}} $\uparrow$ \\ 
    \toprule 
        \textcolor{gray}{\pretrain{}} & \textcolor{gray}{$\times$} & \textcolor{gray}{96.2} & \textcolor{gray}{41.9} & \textcolor{gray}{95.9} & \textcolor{gray}{0.9} & \textcolor{gray}{24.2} & \textcolor{gray}{40.6} & \textcolor{gray}{-} \\ 
        \textcolor{gray}{\retrain{}} & \textcolor{gray}{$\times$} & \textcolor{gray}{96.5} & \textcolor{gray}{31.3} & \textcolor{gray}{95.9} & \textcolor{gray}{1.6} & \textcolor{gray}{27.0} & \textcolor{gray}{34.4} & \textcolor{gray}{-} \\ 
        \textcolor{gray}{\retrain{} + \textsc{gDRO}} & \textcolor{gray}{$\times$} & \textcolor{gray}{95.8} & \textcolor{gray}{67.4} & \textcolor{gray}{95.1} & \textcolor{gray}{13.2} & \textcolor{gray}{12.5} & \textcolor{gray}{63.1} & \textcolor{gray}{-} \\ 
        \retrain{} & \checkmark & 96.3 & 39.7 & 95.8 & 2.2 & 23.9 & 41.3 & - \\ 
    \cmidrule(lr){1-9} 
        \sparse{} & $\times$ & 95.7~(0.5) & 29.0~(10.7) & 95.4~(0.3) & \textbf{1.5}~\textbf{(1.3)} & 28.5~(4.6) & 30.4~(10.9) & 95.3 \\ 
        \salun{} & $\times$ & \textbf{96.2}~\textbf{(0.2)} & 29.3~(10.3) & \textbf{95.8}~\textbf{(0.1)} & 0.7~(1.5) & 29.1~(5.2) & 30.6~(10.7) & 95.3 \\ 
        \scrub{} & $\times$ & 96.5~(0.3) & 35.1~(4.6) & 95.9~(0.2) & 0.6~(1.7) & 26.6~(2.8) & 35.9~(5.4) & 97.5 \\ 
        \rowmethod \method{} & $\times$ & 96.4~(0.3) & \textbf{36.3}~\textbf{(3.4)} & \textbf{95.9}~\textbf{(0.1)} & \textbf{1.0}~\textbf{(1.3)} & 26.1~(2.2) & 36.3~(5.0) & 98.0 \\ 
    \cmidrule(lr){1-9} 
        \sparse{} & \checkmark & 95.6~(0.6) & 37.3~(3.6) & 95.4~(0.4) & 0.7~(1.5) & 26.7~(2.8) & 34.8~(6.5) & 97.4 \\ 
        \salun{} & \checkmark & \textbf{96.1}~\textbf{(0.2)} & 42.9~(8.2) & \textbf{95.8}~\textbf{(0.1)} & 0.6~(2.0) & 23.9~(3.4) & 41.1~(9.8) & 96.1 \\ 
        \scrub{} & \checkmark & 96.4~(0.3) & 43.5~(3.8) & 96.0~(0.2) & 0.7~(1.5) & 23.7~(0.7) & 43.0~(1.7) & 98.6 \\ \rowmethod 
        \method{} & \checkmark & 96.3~(0.3) & 43.2~(3.5) & 96.0~(0.2) & \textbf{1.2}~\textbf{(1.3)} & \textbf{24.0}~\textbf{(0.5)} & \textbf{41.3}~\textbf{(0.4)} & \textbf{99.0} \\
    \end{tabularx}
    \label{tab:robust_celeba}
\end{table*}

\begin{table*}[t]
    \scriptsize
    \centering
    \caption{\textbf{Group-robust machine unlearning in \waterbird{}.} We build the forget set by sampling data points from a single group. The unlearning ratio is set to 0.5. We compare \inlinecolorbox{colormethod}{\method{}} against \sparse{}, \salun{}, and \scrub{}. The \gap{} and deltas are computed against \retrain{} + \reweight{}, and we bold the methods that achieve the smallest discrepancy with it. To avoid confusion, other reference models are in \textcolor{gray}{light gray}.}
    \vspace{\baselineskip}
    \begin{tabularx}{\textwidth} {
        L{4cm}
        C{0.5cm}
        *{6}{Y}
        C{1.5cm}
    }
        method & RW & RA & UA & TA & MIA & EO & GA & \textbf{\gap{}} $\uparrow$ \\
    \toprule
        \textcolor{gray}{\pretrain{}} & \textcolor{gray}{$\times$} & \textcolor{gray}{98.9} & \textcolor{gray}{84.5} & \textcolor{gray}{87.7} & \textcolor{gray}{33.3} & \textcolor{gray}{26.2} & \textcolor{gray}{56.6} & \textcolor{gray}{-} \\
        \textcolor{gray}{\retrain{}} & \textcolor{gray}{$\times$} & \textcolor{gray}{98.7} & \textcolor{gray}{52.4} & \textcolor{gray}{86.5} & \textcolor{gray}{54.8} & \textcolor{gray}{30.4} & \textcolor{gray}{49.4} & \textcolor{gray}{-} \\
        \textcolor{gray}{\retrain{} + \textsc{gDRO}} & \textcolor{gray}{$\times$} & \textcolor{gray}{94.7} & \textcolor{gray}{89.3} & \textcolor{gray}{91.6} & \textcolor{gray}{21.4} & \textcolor{gray}{7.3} & \textcolor{gray}{83.1} & \textcolor{gray}{-} \\
        \retrain{} & \checkmark & 99.0 & 59.5 & 87.2 & 53.6 & 28.3 & 51.6 & - \\
    \cmidrule(lr){1-9}
        \sparse{} & $\times$ & \textbf{99.0~(0.2)} & 59.5~(7.1) & 85.6~(1.6) & 44.0~(9.5) & 32.2~(4.1) & 48.8~(11.1) & 94.4 \\
        \salun{} & $\times$ & 100.0~(1.0) & 50.0~(9.5) & 81.8~(5.4) & 90.5~(36.9) & 38.7~(10.4) & 39.3~(12.3) & 87.4 \\
        \scrub{} & $\times$ & 98.8~(0.3) & 60.7~(10.7) & 86.9~(0.7) & 45.2~(10.7) & 31.9~(4.3) & 41.7~(9.8) & 93.9 \\
        \rowmethod
        \method{} & $\times$ & 100.0~(1.0) & 53.6~(8.3) & 86.1~(1.2) & 58.3~(7.1) & 28.3~(3.0) & 53.8~(7.5) & 95.3 \\
    \cmidrule(lr){1-9}
        \sparse{} & \checkmark & \textbf{98.7~(0.2)} & 64.3~(11.9) & 85.0~(2.2) & 46.4~(7.1) & 30.6~(2.3) & 53.7~(8.3) & 94.7 \\
        \salun{} & \checkmark & 100.0~(1.0) & 47.6~(16.7) & 81.1~(6.1) & 91.7~(38.1) & 39.0~(10.7) & 39.0~(12.5) & 85.8 \\
        \scrub{} & \checkmark & \textbf{98.9~(0.2)} & 66.7~(11.9) & \textbf{87.0~(0.6)} & 44.0~(9.5) & 30.9~(3.4) & 44.3~(7.3) & 94.5 \\
        \rowmethod
        \method{} & \checkmark & 99.9~(0.9) & \textbf{54.8~(4.8)} & 85.8~(1.4) & \textbf{59.5~(6.0)} & \textbf{28.3~(1.8)} & \textbf{53.7~(4.0)} & \textbf{96.9} \\
    \end{tabularx}
    \label{tab:robust_waterbirds}
\end{table*}

\begin{table*}[t]
    \scriptsize
    \centering
    \caption{\textbf{Group-robust machine unlearning in \fairface{}.} We build the forget set by sampling data points from a single group. The unlearning ratio is set to 0.5. We compare \inlinecolorbox{colormethod}{\method{}} against \sparse{}, \salun{}, and \scrub{}. The \gap{} and deltas are computed against \retrain{} + \reweight{}, and we bold the methods that achieve the smallest discrepancy with it. To avoid confusion, other reference models are in \textcolor{gray}{light gray}.}
    \vspace{\baselineskip}
    \begin{tabularx}{\textwidth} {
        L{4cm}
        C{0.5cm}
        *{6}{Y}
        C{1.5cm}
    }
        method & RW & RA & UA & TA & MIA & EO & GA & \textbf{\gap{}} $\uparrow$ \\
    \toprule
        \textcolor{gray}{\pretrain{}} & \textcolor{gray}{$\times$} & \textcolor{gray}{65.6} & \textcolor{gray}{79.0} & \textcolor{gray}{57.2} & \textcolor{gray}{0.2} & \textcolor{gray}{5.4} & \textcolor{gray}{71.2} & \textcolor{gray}{-} \\
        \textcolor{gray}{\retrain{}} & \textcolor{gray}{$\times$} & \textcolor{gray}{66.8} & \textcolor{gray}{57.8} & \textcolor{gray}{56.5} & \textcolor{gray}{0.9} & \textcolor{gray}{9.2} & \textcolor{gray}{58.7} & \textcolor{gray}{-} \\
        \textcolor{gray}{\retrain{} + \textsc{gDRO}} & \textcolor{gray}{$\times$} & \textcolor{gray}{61.7} & \textcolor{gray}{56.3} & \textcolor{gray}{51.4} & \textcolor{gray}{10.2} & \textcolor{gray}{2.3} & \textcolor{gray}{57.4} & \textcolor{gray}{-} \\
        \retrain{} & \checkmark & 66.7 & 69.3 & 56.7 & 0.7 & 5.6 & 69.6 & - \\
    \cmidrule(lr){1-9}
        \sparse{} & $\times$ & 64.0~(2.7) & 74.1~(4.8) & \textbf{56.9~(0.5)} & 0.2~(0.7) & 6.1~(0.9) & 69.4~(0.4) & 98.3 \\
        \salun{} & $\times$ & 66.3~(0.3) & 66.6~(3.0) & 55.9~(0.8) & 0.3~(0.6) & 9.0~(3.4) & 60.3~(9.3) & 97.1 \\
        \scrub{} & $\times$ & 66.9~(0.3) & 65.4~(3.9) & 56.7~(0.6) & 1.0~(0.5) & 9.9~(4.3) & 61.3~(8.3) & 97.0 \\
        \rowmethod
        \method{} & $\times$ & \textbf{66.7~(0.1)} & 74.7~(5.4) & 57.2~(0.8) & 0.3~(0.5) & 6.0~(1.1) & 66.1~(3.5) & 98.1 \\
    \cmidrule(lr){1-9}
        \sparse{} & \checkmark & 64.4~(2.3) & 72.9~(3.6) & 56.0~(0.9) & 0.3~(0.4) & 6.1~(2.0) & 67.0~(7.1) & 97.3 \\
        \salun{} & \checkmark & 65.1~(1.5) & 69.8~(5.6) & 54.8~(1.8) & 0.3~(0.4) & 6.6~(1.7) & 63.7~(5.9) & 97.2 \\
        \scrub{} & \checkmark & 66.7~(0.2) & 73.4~(4.1) & 57.2~(0.6) & \textbf{0.7~(0.3)} & \textbf{6.2~(0.7)} & 70.2~(1.8) & \textbf{98.7} \\
        \rowmethod
        \method{} & \checkmark & 64.7~(1.9) & \textbf{71.6~(2.3)} & \textbf{57.1~(0.5)} & \textbf{0.3~(0.3)} & 5.8~(1.5) & \textbf{70.3~(1.2)} & \textbf{98.7} \\
    \end{tabularx}
    \label{tab:robust_fairface}
\end{table*}

\Cref{tab:robust_celeba,tab:robust_waterbirds,tab:robust_fairface} show results for group-robust unlearning on \celeba{}, \waterbird{}, and \fairface{} using an unlearning ratio $r$ of $0.5$.

\noindent\textbf{CelebA.}
\retrain{} + \reweight{} achieves the best trade-off between original performance preservation and unlearning, showing the best gap with UA (-2.2), EO (-0.3), and GA (+0.7).
The forget accuracy (UA) is very close to that of \pretrain{}, which might seem strange at first glance.
Yet, UA and GA share similar values for both \pretrain{} and \retrain{}. 
This suggests that the model achieves a low generalization error, as the forget accuracy aligns with that of unseen samples (from the same group) \emph{regardless} of whether the forget set was part of the training data.
Additionally, as \celeba{} counts numerous images, \reweight{} easily preserves original group accuracies, thus, recovering the original model performances and showing the same UA of \pretrain{}.
Retraining the model with \dro{} generally leads to unwanted unlearning behaviors, \ie, the UA increases by 25.5 points.
Plain \retrain{}, instead, suffers performance degradation in the dominant group of the forget set, as we highlight in \cref{fig:ratio} (GA lowers by 6.2 points).
Adopting the proposed reweighting strategy leads to a better trade-off in group-robust unlearning.

\begin{wrapfigure}[22]{r}{0.5\linewidth}
    \vspace{-\baselineskip}
    \centering
    \includegraphics[width=\linewidth]{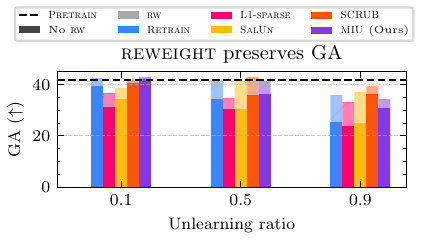}
    \caption{\changed{\textbf{\reweight{} for group-robust unlearning.} As in \cref{fig:ratio}, we test different methods and \reweight{} in group-robust unlearning on CelebA. Darker colors are used for methods without the reweighting, while lighter ones correspond to methods coupled with \reweight{}.
    As the unlearning ratio grows, the methods' GA degrade. Instead, adding \reweight{} restores the original GA.}}
    \label{fig:reweight}
\end{wrapfigure}

\method{} achieves the best performance preservation (\cref{tab:robust_celeba}) by scoring the highest GA alignment both when using~($\Delta$0.4) and not using~($\Delta$5.0) \reweight{}.
These results highlight that mutual information improves group performance retention.
Instead, existing approaches must rely on \reweight{} to close the gap in GA and EO with \method{} as they are agnostic to group-robust machine unlearning.
We also notice that using \reweight{} increases the UA for all tested algorithms.
Yet, despite the method used, coupling \reweight{} with approximate unlearning always recovers most of the original GA.
We visualize this in \cref{fig:reweight}, which shows the same experiment as \cref{fig:ratio} while highlighting the \reweight{} contribution.
However, as \method{} is strictly designed for this task, it generally achieves better robustness than existing approaches, scoring the best \gap{}.

\noindent\textbf{Waterbirds.}
Similarly to \celeba{} experiments, \retrain{} + \reweight{} achieves the best discrepancy (\cref{tab:robust_waterbirds}) in terms of TA~(-0.5), EO~(+2.1), and GA~(-5.0), better preserving original group robustness.
\retrain{} + \dro{}, instead, increases the UA by 4.8\% above \pretrain{} and by 36.9\% above \retrain{}, which can negatively affect the unlearning evaluation if used as the gold standard for approximate machine unlearning.
Compared to the \celeba{} case, \retrain{} achieves a better unlearning-preservation trade-off.
Nonetheless, \retrain{} + \reweight{} is still a better candidate for group-robust unlearning, always achieving the best calibration with the original model.

In \waterbird{}, our method outperforms existing methods both with and without \reweight{}.
Yet, \reweight{} does not provide a substantial improvement as only the forget accuracy~(from a delta of 8.3 to 4.8) and the EO~(from a delta of 7.5 to 4.0) get significantly enhanced.
Instead, \reweight{} substantially increases \sparse{} and \scrub{} average UA (+4.8 and +6.0), achieving higher values than \retrain{} + \reweight{}~(59.5\%).
Nonetheless, it also improves the GA, showing better dominant forget group preservation.
We argue that the subtle improvement provided by \reweight{} is caused by the limited dataset size of \waterbird{}.
By increasing the sampling likelihood of the few dominant forget group images left, the network overfits those few samples, limiting robustness preservation.
Even without \reweight{}, \method{} outperforms all baselines, regardless of whether they use \reweight{}, further validating our design choices.

\noindent\textbf{FairFace.}
Differently from the other datasets, here \retrain{} + \dro{} struggles to preserve original model accuracy (\cref{tab:robust_fairface}) in both the dominant forget group~(-13.8) and the test set~(-5.8).
We ascribe this behavior to numerous \fairface{} groups that make the group-robust optimization challenging.
Instead, \retrain{} + \reweight{} overcomes this issue as it simply reweights group sampling probabilities to match the original training dataset statistics, achieving better GA (-1.6) and a TA (-0.5).
Importantly, our experiments highlight a key advantage of \reweight{}, which functions effectively ``off the shelf'' with the original model hyperparameters. 
Unlike \retrain{} + \dro{}, it requires \emph{no additional tuning}.

All methods show good alignment to \retrain{} + \reweight{}, even without reweighting.
TA and GA are generally preserved, with \salun{} scoring lowest at 55.9\% (delta 0.8) and 60.3\% (delta 9.3) respectively.
However, methods without \reweight{} show higher UA than \retrain{}, indicating partial scrubbing of the forget set.
Regardless, \cref{sub:ablations} shows that the high UA of \method{} is not caused by poor unlearning but the \emph{calibration term}, which recovers the original group robustness.
\reweight{} further enhances model robustness, reflected in higher GA and lower EO.
Finally, \method{} + \reweight{} and \scrub{} + \reweight{} achieve the same \gap{}, but while our method shows a better UA and GA alignment with \retrain{} + \reweight{}, \scrub{} obtains a better alignment in EO terms.
Although \reweight{} does not fundamentally improve the \gap{}, it generally enhances GA and EO, promoting performance retention.

\noindent\textbf{Discussion.} 
UA strongly correlates with GA after unlearning in all three datasets. 
A high UA-GA correlation suggests that the forget set behaves as unseen data of the same group, indicating that the forget set was \emph{properly} unlearned.
Moreover, \method{} consistently approximates \retrain{}+\reweight{} better than existing methods or is on par, while \reweight{} reliably preserves performance without drawbacks in model retraining or approximate unlearning.

\subsection{Impact of different unlearning ratios}
\label{sub:varying}
\begin{figure}
    \begin{minipage}[t]{0.48\linewidth}
        \centering
        \includegraphics[width=\linewidth]{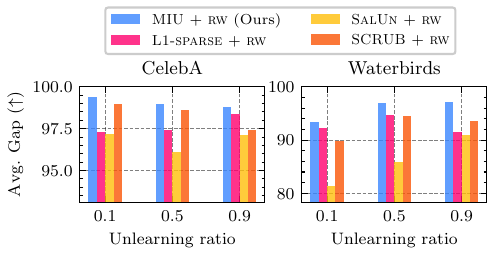}
        \caption{\changed{\textbf{Group-robust unlearning across different unlearning ratios.} 
        We compare \sparse{}, \salun{}, and \scrub{} against our approach while using the \reweight{} strategy on all methods. \method{} achieves overall the best \gap{} when varying the unlearning ratio.}}
        \label{fig:robust_ratio}
    \end{minipage}
    \hfill
    \begin{minipage}[t]{0.48\textwidth}
        \centering
        \includegraphics[width=\linewidth]{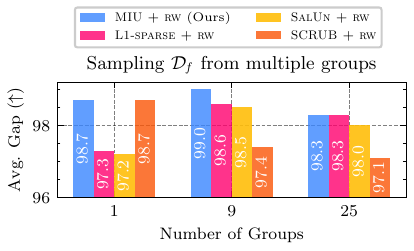}
        \caption{\textbf{Sampling the forget set from multiple groups.} We evaluate our method against \sparse{}, \salun{}, and \scrub{} when the forget set is sampled from multiple \fairface{} groups. \method{} is more consistent across experiments, always achieving the best result.}
        \label{fig:multi_group}
    \end{minipage}
\end{figure}
\Cref{fig:robust_ratio} analyzes methods \gap{} across different unlearning ratios.
On \celeba{}, all methods show consistent discrepancies from \retrain{} + \reweight{} across different unlearning ratios, likely because \reweight{} recovers a large portion of the lost robustness.
Nonetheless, \method{} consistently outperforms existing approaches, scoring 99.4\%, 99.0\%, and 98.8\% at unlearning ratios of 0.1, 0.5, and 0.9.
Instead, all algorithms struggle at 0.1 unlearning ratio in \waterbird{}, \changed{where the small forget set size (\ie, five samples) causes high fluctuations in the UA, as each sample counts as $\nicefrac{1}{5}$ of the total classification error.
Furthermore, the reduced forget set size makes the gradient and BatchNorm \citep{ioffe2015batch} estimation noisy.}
As the unlearning ratio grows (\ie, 0.5 and 0.9), \method{} outperforms baselines by a growing margin.
At unlearning ratio 0.9, \method{} achieves 97.2\% on Waterbirds (\vs{} 93.6\% of \scrub{}).
Furthermore, \method{} remains more consistent across unlearning ratios, confirming that our design choices effectively narrow the \gap{} with \retrain{} + \reweight{}. 
Full~\cref{fig:robust_ratio} tables are reported in \cref{appx:additional_results}.

\subsection{Multi-group unlearning}
\label{sub:multigroup}
This section compares \method{} and existing approaches in multi-group unlearning, \ie, when the forget set data is sampled from multiple groups \changed{(group composition is in \cref{appxsub:groups})}.
\Cref{fig:multi_group} shows the outcome of this experiment on \fairface{}, given its numerous groups, with a varying number of groups in the forget set and an unlearning ratio fixed to 0.5.
All methods achieve high scores and a good discrepancy with \retrain{} + \reweight{}, though \scrub{} performs worst (97.1\% of \gap{}).
Performance gaps shrink as more groups are included, as forget-set statistics align with the original training distribution, reducing non-uniform sampling effects.
Nonetheless, \method{} is the most consistent, highlighting its effectiveness in group-robust machine unlearning.

\subsection{Ablations}
\label{sub:ablations}
\Cref{tab:ablations} shows the ablation of \method{} components in all three datasets by systematically adding each element to understand its contribution.
We mark with ``\checkmark{}'' when the component is used in the experiment and ``$\times$'' when it is not.
From left to right, we list \emph{retaining term}, \emph{unlearning term}, \emph{calibration term}, and \reweight{}.
In the first row, we consider our method when only the \emph{unlearning term} and the \emph{retaining term} are used.
We highlight how the UA is low for all three datasets, demonstrating that mutual information minimization can be used to unlearn.
\begin{wraptable}[19]{r}{0.55\textwidth}
    \vspace{-\baselineskip}
    \scriptsize
    \centering
    \caption{\changed{\textbf{\method{} ablations.} We compute \method{} ablations on each of the three investigated datasets. From left to right, we report the investigated dataset, \emph{retaining term}, \emph{unlearning term}, \emph{calibration term}, and \reweight{}. We measure performance using UA, GA, and \gap{}. The configuration that corresponds to \inlinecolorbox{colormethod}{\method{} + \reweight{}} is highlighted.}}
    \vspace{\baselineskip}
    \begin{tabularx}{\linewidth} { 
        L{1.4cm}
        *{6}{Y}
        C{0.7cm}
    }
            dataset & \cref{eq:retain} & \cref{eq:unlearn} & \cref{eq:regularize} & RW & UA & GA & gap $\uparrow$ \\
        \toprule
            \multirow{4}{*}{CelebA}
             & \checkmark & \checkmark & $\times$ & $\times$ & 35.5 & 35.4 & 97.5 \\
             & \checkmark & \checkmark & \checkmark & $\times$ & 36.3 & 36.3 & 98.0 \\
             & $\times$ & \checkmark & \checkmark & $\times$ & 27.2 & 28.5 & 95.2 \\
             & \cellcolor{colormethod}\checkmark & \cellcolor{colormethod}\checkmark & \cellcolor{colormethod}\checkmark & \cellcolor{colormethod}\checkmark & \cellcolor{colormethod}43.2 & \cellcolor{colormethod}41.3 & \cellcolor{colormethod}99.0 \\
        \cmidrule{1-8}
            \multirow{4}{*}{Waterbirds}
             & \checkmark & \checkmark & $\times$ & $\times$ & 47.6 & 51.1 & 92.5 \\
             & \checkmark & \checkmark & \checkmark & $\times$ & 53.6 & 53.8 & 95.3 \\
             & $\times$ & \checkmark & \checkmark & $\times$ & 16.7 & 16.8 & 81.9 \\
             & \cellcolor{colormethod}\checkmark & \cellcolor{colormethod}\checkmark & \cellcolor{colormethod}\checkmark & \cellcolor{colormethod}\checkmark & \cellcolor{colormethod}54.8 & \cellcolor{colormethod}53.7 & \cellcolor{colormethod}96.9 \\
        \cmidrule{1-8}
            \multirow{4}{*}{FairFace}
             & \checkmark & \checkmark & $\times$ & $\times$ & 63.1 & 59.2 & 96.1 \\
             & \checkmark & \checkmark & \checkmark & $\times$ & 74.7 & 66.1 & 98.1 \\
             & $\times$ & \checkmark & \checkmark & $\times$ & 87.1 & 81.1 & 93.0 \\
             & \cellcolor{colormethod}\checkmark & \cellcolor{colormethod}\checkmark & \cellcolor{colormethod}\checkmark & \cellcolor{colormethod}\checkmark & \cellcolor{colormethod}71.6 & \cellcolor{colormethod}70.3 & \cellcolor{colormethod}98.7 \\
    \end{tabularx}
    \label{tab:ablations}
\end{wraptable}
 
This baseline already achieves a remarkable \gap{} with \retrain{} + \reweight{}, scoring a 97.5\% in \celeba{}, which is the best among methods that do not use \reweight{}.
Adding our \emph{calibration term} (\cref{eq:regularize}), leads to an increase in GA in all three datasets, with \fairface{} showing a growth of 6.9\%.
We highlight that the forget accuracy also grows.
Yet, this increase is caused by \cref{eq:regularize}, which calibrates the mutual information to match the original group robustness.
Therefore, the high UA cannot be blamed on the poor unlearning.
Although most previous approaches exploit a \emph{retaining term} \citep{chundawat2023can,kurmanji2024towards,jia2024modelsparsitysimplifymachine,fan2023salun}, we also ablate this component for completeness.
As previous works suggest \citep{kurmanji2024towards,chundawat2023can}, removing the \emph{retaining term} negatively impacts model utility, resulting in the lowest performance overall (\eg, 81.9\% in Waterbirds).
Similarly, when adding \reweight{}, the \gap{} gets improved in all three datasets (\eg, 96.9\% in Waterbirds), as we also highlight in \cref{fig:reweight}.
These results highlight the contribution of each of \method{} components that allow for unlearning (\cref{eq:unlearn}) while preserving forget set dominant group performance (\cref{eq:regularize}).

\section{Conclusion}
\label{sec:conclusion}
This paper is the first to address the performance degradation \changed{that can be} caused by non-uniformly distributed forget sets in both model retraining and approximate unlearning.
We show that adopting a simple data distribution reweighting (\reweight{}) for retraining the model is a simple and better alternative than retraining with \dro{}.
Moreover, we propose the first approximate unlearning method (\method{}) that unlearns personal data while reducing the risk of degradation of the forget set dominant group.
Our evaluation demonstrates that \retrain{} + \reweight{} consistently improves over a simple \retrain{} while \method{} outperforms existing baselines in group-robust machine unlearning.

\noindent\textbf{Limitations.}
One limitation of our work is the assumption that group annotations are known, which may not hold in real-world applications where such labels are difficult to obtain.
Therefore, a natural follow-up of this work is a group-agnostic methodology that preserves model robustness as achieved by \reweight{}.
\changed{A na\"ive approach towards this direction would be to discover groups from data \citep{kim2024discovering,d2024openbias}, and applying approaches presented in this paper.}
Furthermore, our evaluation is restricted to the classification setting.
Applying the proposed techniques and baselines to other domains might be non-trivial, and there is no guarantee that unlearning effectiveness and accuracy will transfer directly. 
Despite these limitations, our work is an important first step toward understanding and mitigating the accuracy degradation caused by group-unbalanced forget sets.

\noindent\textbf{Broader impact statement.}
Machine unlearning is designed to remove user data from a trained model.
While its primary goal is to preserve privacy, it can also be misused for malicious purposes.
Targeted unlearning of specific groups may lead to biased models with harmful consequences.
However, the proposed \emph{group-robust machine unlearning} seeks to minimize performance degradation for dominant groups in the forget set.
Thus, \method{} and \reweight{} counteract biases introduced by unlearning, mitigating the negative societal impacts of its misuse.

\noindent\textbf{Acknowledgements.}
We acknowledge the CINECA award under the ISCRA initiative for the availability of high-performance computing resources and support. Elisa Ricci and Massimiliano Mancini are supported by the MUR PNRR project FAIR - Future AI Research (PE00000013), funded by NextGeneration EU. Elisa Ricci is also supported by the EU projects AI4TRUST (No.101070190) and ELIAS (No.01120237). Thomas De Min is funded by NextGeneration EU. This work is also supported by the EU project ELLIOT (101214398) and by the French National Research Agency (ANR) with the ANR-20-CE23-0027. We thank Olivier Laurent for his valuable feedback and insightful suggestions.

\bibliography{main}
\bibliographystyle{tmlr}

\clearpage
\appendix
\section{Metrics}
\label{appx:metrics}
This section provides further details on the adopted metrics, \ie, the RA, UA, TA, the membership-inference attack (MIA) \citep{yeom2018privacy}, the dominant forget group accuracy (GA), the equalized odds (EO), and the average gap (\gap{}) \citep{fan2023salun}. 

\noindent\textbf{RA, UA, and TA.}
We evaluate the model on the remaining, forget, and test sets to compute the remaining, forget, and test accuracy.
Therefore, we report the ratio of correctly classified samples for each of these subsets of the dataset.

\noindent\textbf{MIA.}
To compute the membership inference attack, we follow previous works \citep{jia2024modelsparsitysimplifymachine,fan2023salun} and train a model on remaining and validation losses to predict membership.
Therefore, we assign a different binary label to remaining and validation losses and train the model to discriminate between them.
After model training, we use such a model to infer the membership of forget set data.
In all tables, we report the MIA-Efficacy \citep{jia2024modelsparsitysimplifymachine} that is computed as follows:
\begin{equation}
    \text{MIA-Efficacy} = \frac{\mathit{TN}}{|\forgetds|}\text{,}
\end{equation}
where $\mathit{TN}$ are the true negatives, \ie, the number of samples the MIA predicted as non-members.
Instead of training a support vector machine as in \citep{jia2024modelsparsitysimplifymachine,fan2023salun}, we used a random forest as the accuracies are comparable with an SVM.
Moreover, training is faster since it can be easily parallelized.
The higher the MIA-Efficacy, the better the model's privacy protection.

\noindent\textbf{EO.}
Equalized Odds \citep{hardt2016equality} measures model fairness or prediction dependencies on protected attributes.
To compute the EO, we measure model performance discrepancy by varying the sensitive attribute value and averaging over different target labels.
Formally, the EO for a binary classification model is computed as follows \citep{hardt2016equality}:
\begin{equation}
    \begin{aligned}
    \text{EO} = \frac{1}{2}\sum_{y=0}^1 \left| \p(\hat{\targets}=1\mid \targets=y, \attributes=0)\, -\, \p(\hat{\targets}=1\mid \targets=y, \attributes=1) \right| \text{,}
    \end{aligned}
\end{equation}
where $\hat{\targets}, \targets, \attributes$ are random variables describing model predictions, target attributes, and sensitive attributes.
EO measures the absolute difference in outputting a positive prediction when the protected attribute equals 1 and 0, averaging over the two target attributes.
In the \fairface{} case, we set all classes and protected attributes that do not match those of the dominant group of the forget set as $y=0$ and $a=0$, as FairFace counts more than two classes and more than two attributes.
The lower the EO, the better the model fairness.

\noindent\textbf{\gap{}.} 
Following previous works \citep{jia2024modelsparsitysimplifymachine,fan2023salun}, we compute the \gap{} to simplify the comparison among different methodologies.
\gap{} is computed as the average metric discrepancy between the unlearned model and the retrained gold standard.
Formally, let $M = \{\text{RA},\text{UA},\text{TA},\text{MIA},\text{EO},\text{GA}\}$ be the set of all metrics used in this paper, then \gap{} is computed as follows:
\begin{equation}
    \text{\gap{}} = \overline{\sum}_{m \in M} 1 - \left| \overline{\sum}_{s\in S} m(\varphi_u^s, \theta_u^s) - m(\varphi_r^s, \theta_r^s) \right| \text{,}
\end{equation}
where $m(\varphi_u^s, \theta_u^s)$ and $m(\varphi_r^s, \theta_r^s)$ are calculated using unlearned and retrained model weights, $s$ is the experiment seed and $\overline{\sum}$ is the average.
The closer \gap{} is to 1 (or 100 in the tables), the better the approximation of the retrained model.

\section{Additional results}
\label{appx:additional_results}
This section contains the additional results that could not be included in the main paper.
\Cref{appxsub:non_uniform} intuitively shows the accuracy degradation caused by non-uniformly sampled forget sets in all three investigated datasets.
\Cref{appxsub:ratios,appxsub:groups} report tables associated with experiments on different unlearning ratios and sampling the forget set from multiple groups.
\Cref{appxsub:ablations} further expands \method{} ablations.
Finally, \Cref{appxsub:fairness_metrics} further evaluates \method{}'s and \reweight{}'s fairness and robustness preservation.

\begin{figure}[tp]
    \centering
    \includegraphics[width=\linewidth]{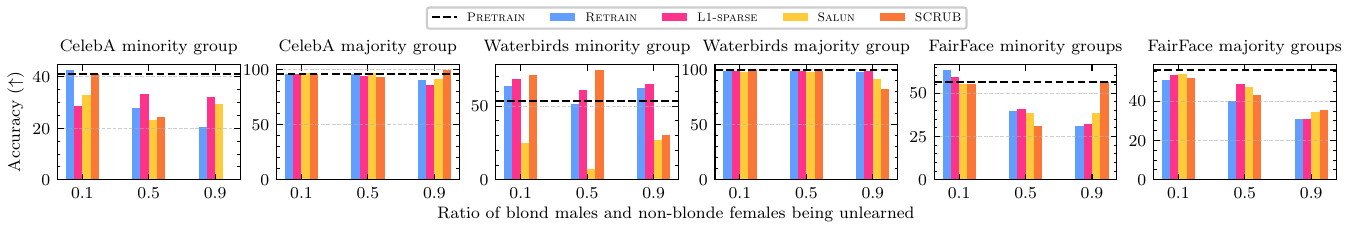}
    \caption{\changed{\textbf{Unlearning non-uniformly sampled data.} We test standard model retraining, and popular approximate unlearning methods (\sparse{}, \salun{}, \scrub{}) in group-robust unlearning. The more samples of least represented groups are unlearned, the lower the model accuracy on such groups. On the contrary, the most represented ones are less affected.}}
    \label{fig:ratio_full}
\end{figure}
\subsection{Unlearning non-uniformly sampled data}
\label{appxsub:non_uniform}
\Cref{fig:ratio} of the main paper shows how the forget set dominant group accuracy drops for least represented groups when increasing the unlearning ratio, limiting the analysis to \celeba{} due to space constraints.
This section reports the same experiment on all three investigated datasets for completeness.

\Cref{fig:ratio_full} shows accuracy variations when unlearning \emph{\targetattr{blond} \sensitiveattr{males}} (minority) and \emph{\targetattr{non-blonde} \sensitiveattr{females}} (majority) in \celeba{}, \emph{\targetattr{waterbirds} on \sensitiveattr{land}} (minority) and \emph{\targetattr{landbirds} on \sensitiveattr{land}} (majority) in \waterbird{}, and \emph{\targetattr{more than 70 y.o.\ }\sensitiveattr{Middle Easterns}} (minority), \emph{\targetattr{60-69 y.o.\ }\sensitiveattr{Caucasians}} (minority), \emph{\targetattr{20-29 y.o.\ }\sensitiveattr{African Americans}} (majority), and \emph{\targetattr{30-39 y.o.\ }\sensitiveattr{Southeast Asians}} (majority) in \fairface{}, with different unlearning ratios: $0.1$, $0.5$, $0.9$.
While the accuracy drop is more evident on \celeba{}, it is also visible in the other two datasets.
Interestingly, some unlearning methods are more robust to this performance drop than others (\eg, \scrub{} scores a 56.25\% in \fairface{} with unlearning ratio 0.9 \vs{} 38.54\% of \salun{} in minority groups, where the latter is the second best). 

Although some methods are more robust, we argue that the quality of the unlearning process influences the accuracy of the dominant group of the forget set.
The less the forget set was unlearned, the more the performance retention.
As an example, \cref{sub:results} highlights that likely \scrub{} fails to effectively unlearn the forget set in the \fairface{} experiment, showing higher minority group accuracy (56.25\%) than the \retrain{} (31.25\%).
Thus, to better investigate the relationship between unlearning effectiveness and performance retention, \cref{appxsub:ratios} reports tables associated with \cref{fig:ratio_full} using all investigated metrics.

\begin{table}[tp]
    \scriptsize
    \centering
    \caption{\changed{\textbf{Group-robust machine unlearning in \celeba{} with 0.1 unlearning ratio.}
    We build the forget set by sampling data points from a single group. The unlearning ratio is set to 0.1. We compare \inlinecolorbox{colormethod}{\method{}} against \sparse{}, \salun{}, and \scrub{}. The \gap{} is computed against \retrain{} + \reweight{}.}}
    \vspace{\baselineskip}
    \begin{tabularx}{\textwidth} {
        L{4cm}
        C{0.5cm}
        *{6}{Y}
        C{1.5cm}
    }
        method & RW & RA & UA & TA & MIA & EO & GA & \textbf{\gap{}} $\uparrow$ \\
    \toprule
        \pretrain{} & $\times$ & 96.2\scriptsize$\pm$0.0  & 44.0\scriptsize$\pm$2.7  & 95.9\scriptsize$\pm$0.1  & 0.5\scriptsize$\pm$0.7  & 23.4\scriptsize$\pm$0.5  & 41.9\scriptsize$\pm$1.7  & - \\
        \retrain{} & $\times$ & 96.2\scriptsize$\pm$0.0  & 39.9\scriptsize$\pm$1.0  & 95.8\scriptsize$\pm$0.0  & 0.2\scriptsize$\pm$0.3  & 24.6\scriptsize$\pm$0.3  & 39.6\scriptsize$\pm$0.7  & - \\
        \retrain{} & \checkmark & 96.2\scriptsize$\pm$0.0  & 44.4\scriptsize$\pm$2.1  & 95.9\scriptsize$\pm$0.0  & 0.7\scriptsize$\pm$0.6  & 23.4\scriptsize$\pm$0.7  & 42.4\scriptsize$\pm$2.2  & - \\
    \cmidrule(lr){1-9}
        \sparse{} & $\times$ & 95.4\scriptsize$\pm$0.1& 36.0\scriptsize$\pm$3.9& 95.4\scriptsize$\pm$0.1& 0.7\scriptsize$\pm$0.0& 27.6\scriptsize$\pm$0.7& 31.3\scriptsize$\pm$4.4& 95.7\scriptsize$\pm$1.2\\
        \salun{} & $\times$ & 96.2\scriptsize$\pm$0.0& 36.2\scriptsize$\pm$1.0& 95.8\scriptsize$\pm$0.0& 0.0\scriptsize$\pm$0.0& 26.9\scriptsize$\pm$0.6& 34.4\scriptsize$\pm$2.3& 96.6\scriptsize$\pm$1.4\\
        \scrub{} & $\times$ & 96.4\scriptsize$\pm$0.0& 42.8\scriptsize$\pm$2.1& 96.0\scriptsize$\pm$0.0& 0.5\scriptsize$\pm$0.7& 24.5\scriptsize$\pm$0.3& 40.6\scriptsize$\pm$0.8& 98.8\scriptsize$\pm$0.8\\
        \rowmethod
        \method{} & $\times$ & 96.2\scriptsize$\pm$0.0& 43.0\scriptsize$\pm$2.4& 95.9\scriptsize$\pm$0.0& 0.2\scriptsize$\pm$0.3& 23.4\scriptsize$\pm$0.1& 42.2\scriptsize$\pm$0.5& 99.1\scriptsize$\pm$0.3\\
    \cmidrule(lr){1-9}
        \sparse{} & \checkmark & 95.4\scriptsize$\pm$0.0& 38.9\scriptsize$\pm$1.9& 95.4\scriptsize$\pm$0.0& 1.0\scriptsize$\pm$0.3& 26.3\scriptsize$\pm$1.4& 36.5\scriptsize$\pm$4.1& 97.3\scriptsize$\pm$1.0\\
        \salun{} & \checkmark & 96.1\scriptsize$\pm$0.0& 40.1\scriptsize$\pm$5.9& 95.9\scriptsize$\pm$0.0& 0.7\scriptsize$\pm$0.6& 25.0\scriptsize$\pm$1.9& 38.7\scriptsize$\pm$6.1& 97.2\scriptsize$\pm$0.9\\
        \scrub{} & \checkmark & 96.4\scriptsize$\pm$0.0& 44.7\scriptsize$\pm$3.0& 96.0\scriptsize$\pm$0.0& 0.5\scriptsize$\pm$0.3& 23.9\scriptsize$\pm$0.4& 42.0\scriptsize$\pm$1.3& 99.0\scriptsize$\pm$0.4\\
        \rowmethod
        \method{} & \checkmark & 96.2\scriptsize$\pm$0.0& 44.9\scriptsize$\pm$2.7& 95.9\scriptsize$\pm$0.0& 0.2\scriptsize$\pm$0.3& 22.9\scriptsize$\pm$0.5& 43.0\scriptsize$\pm$0.7& 99.4\scriptsize$\pm$0.1\\
    \end{tabularx}
    \label{tab:celeba_0.1}
\end{table}

\begin{table*}[t]
    \scriptsize
    \centering
    \caption{\changed{\textbf{Group-robust machine unlearning in \celeba{} with 0.5 unlearning ratio.} We build the forget set by sampling data points from a single group. The unlearning ratio is set to 0.5. We compare \inlinecolorbox{colormethod}{\method{}} against \sparse{}, \salun{}, and \scrub{}. The \gap{} is computed against \retrain{} + \reweight{}.}}
    \vspace{\baselineskip}
    \begin{tabularx}{\textwidth} {
        L{4cm}
        C{0.5cm}
        *{6}{Y}
        C{1.5cm}
    }
        method & RW & RA & UA & TA & MIA & EO & GA & \textbf{\gap{}} $\uparrow$ \\
    \toprule
        \pretrain{} & $\times$ & 96.2\scriptsize$\pm$0.2  & 41.9\scriptsize$\pm$0.8  & 95.9\scriptsize$\pm$0.1  & 0.9\scriptsize$\pm$0.3  & 24.2\scriptsize$\pm$1.0  & 40.6\scriptsize$\pm$2.1  & - \\
        \retrain{} & $\times$ & 96.5\scriptsize$\pm$0.0  & 31.3\scriptsize$\pm$0.3  & 95.9\scriptsize$\pm$0.0  & 1.6\scriptsize$\pm$0.4  & 27.0\scriptsize$\pm$0.4  & 34.4\scriptsize$\pm$0.9  & - \\
        \retrain{} & \checkmark & 96.3\scriptsize$\pm$0.1  & 39.7\scriptsize$\pm$0.7  & 95.8\scriptsize$\pm$0.0  & 2.2\scriptsize$\pm$1.2  & 23.9\scriptsize$\pm$0.8  & 41.3\scriptsize$\pm$1.6  & - \\
    \cmidrule(lr){1-9}
        \sparse{} & $\times$ & 95.7\scriptsize$\pm$0.1& 29.0\scriptsize$\pm$3.1& 95.4\scriptsize$\pm$0.1& 1.5\scriptsize$\pm$0.6& 28.5\scriptsize$\pm$0.6& 30.4\scriptsize$\pm$2.9& 95.3\scriptsize$\pm$0.7\\
        \salun{} & $\times$ & 96.2\scriptsize$\pm$0.1& 29.3\scriptsize$\pm$7.7& 95.8\scriptsize$\pm$0.1& 0.7\scriptsize$\pm$0.2& 29.1\scriptsize$\pm$1.5& 30.6\scriptsize$\pm$7.5& 95.3\scriptsize$\pm$2.5\\
        \scrub{} & $\times$ & 96.5\scriptsize$\pm$0.2& 35.1\scriptsize$\pm$1.7& 95.9\scriptsize$\pm$0.0& 0.6\scriptsize$\pm$0.2& 26.6\scriptsize$\pm$1.1& 35.9\scriptsize$\pm$2.0& 97.5\scriptsize$\pm$0.4\\
        \rowmethod
        \method{} & $\times$ & 96.4\scriptsize$\pm$0.1& 36.3\scriptsize$\pm$0.9& 95.9\scriptsize$\pm$0.1& 1.0\scriptsize$\pm$0.3& 26.1\scriptsize$\pm$0.8& 36.3\scriptsize$\pm$2.0& 98.0\scriptsize$\pm$0.4\\
    \cmidrule(lr){1-9}
        \sparse{} & \checkmark & 95.6\scriptsize$\pm$0.0& 37.3\scriptsize$\pm$3.8& 95.4\scriptsize$\pm$0.1& 0.7\scriptsize$\pm$0.3& 26.7\scriptsize$\pm$0.9& 34.8\scriptsize$\pm$4.8& 97.4\scriptsize$\pm$1.4\\
        \salun{} & \checkmark & 96.1\scriptsize$\pm$0.1& 42.9\scriptsize$\pm$10.5& 95.8\scriptsize$\pm$0.0& 0.6\scriptsize$\pm$0.5& 23.9\scriptsize$\pm$2.8& 41.1\scriptsize$\pm$9.8& 96.1\scriptsize$\pm$2.0\\
        \scrub{} & \checkmark & 96.4\scriptsize$\pm$0.2& 43.5\scriptsize$\pm$0.1& 96.0\scriptsize$\pm$0.1& 0.7\scriptsize$\pm$0.1& 23.7\scriptsize$\pm$0.7& 43.0\scriptsize$\pm$0.9& 98.6\scriptsize$\pm$0.3\\
        \rowmethod
        \method{} & \checkmark & 96.3\scriptsize$\pm$0.2& 43.2\scriptsize$\pm$0.6& 96.0\scriptsize$\pm$0.0& 1.2\scriptsize$\pm$0.1& 24.0\scriptsize$\pm$0.5& 41.3\scriptsize$\pm$1.1& 99.0\scriptsize$\pm$0.2\\
    \end{tabularx}
    \label{tab:celeba_0.5}
\end{table*}

\begin{table*}[t]
    \scriptsize
    \centering
    \caption{\changed{\textbf{Group-robust machine unlearning in \celeba{} with 0.9 unlearning ratio.}
    We build the forget set by sampling data points from a single group. The unlearning ratio is set to 0.9. We compare \inlinecolorbox{colormethod}{\method{}} against \sparse{}, \salun{}, and \scrub{}. The \gap{} is computed against \retrain{} + \reweight{}.}}
    \vspace{\baselineskip}
    \begin{tabularx}{\textwidth} {
        L{4cm}
        C{0.5cm}
        *{6}{Y}
        C{1.5cm}
    }
        method & RW & RA & UA & TA & MIA & EO & GA & \textbf{\gap{}} $\uparrow$ \\
    \toprule
        \pretrain{} & $\times$ & 96.4\scriptsize$\pm$0.1  & 44.0\scriptsize$\pm$6.6  & 95.9\scriptsize$\pm$0.1  & 0.7\scriptsize$\pm$0.1  & 23.5\scriptsize$\pm$1.3  & 41.1\scriptsize$\pm$5.2  & - \\
        \retrain{} & $\times$ & 96.7\scriptsize$\pm$0.0  & 20.2\scriptsize$\pm$1.0  & 95.9\scriptsize$\pm$0.0  & 7.2\scriptsize$\pm$1.4  & 31.5\scriptsize$\pm$0.6  & 25.4\scriptsize$\pm$1.9  & - \\
        \retrain{} & \checkmark & 96.4\scriptsize$\pm$0.1  & 33.6\scriptsize$\pm$1.1  & 95.7\scriptsize$\pm$0.1  & 4.3\scriptsize$\pm$1.3  & 25.6\scriptsize$\pm$0.7  & 35.9\scriptsize$\pm$1.9  & - \\
    \cmidrule(lr){1-9}
        \sparse{} & $\times$ & 95.9\scriptsize$\pm$0.1& 21.6\scriptsize$\pm$6.6& 95.4\scriptsize$\pm$0.1& 3.9\scriptsize$\pm$1.0& 31.5\scriptsize$\pm$1.6& 23.9\scriptsize$\pm$7.1& 94.7\scriptsize$\pm$2.6\\
        \salun{} & $\times$ & 96.6\scriptsize$\pm$0.1& 20.8\scriptsize$\pm$0.8& 95.9\scriptsize$\pm$0.0& 2.5\scriptsize$\pm$0.8& 31.5\scriptsize$\pm$0.6& 25.0\scriptsize$\pm$2.4& 94.6\scriptsize$\pm$0.8\\
        \scrub{} & $\times$ & 96.7\scriptsize$\pm$0.1& 27.2\scriptsize$\pm$1.2& 95.9\scriptsize$\pm$0.0& 3.1\scriptsize$\pm$0.5& 31.3\scriptsize$\pm$0.4& 26.3\scriptsize$\pm$0.7& 96.0\scriptsize$\pm$0.6\\
        \rowmethod
        \method{} & $\times$ & 96.5\scriptsize$\pm$0.1& 32.6\scriptsize$\pm$0.9& 95.8\scriptsize$\pm$0.1& 1.5\scriptsize$\pm$0.2& 27.9\scriptsize$\pm$0.3& 30.9\scriptsize$\pm$1.1& 98.1\scriptsize$\pm$0.7\\
    \cmidrule(lr){1-9}
        \sparse{} & \checkmark & 95.9\scriptsize$\pm$0.0& 33.2\scriptsize$\pm$2.9& 95.2\scriptsize$\pm$0.1& 2.6\scriptsize$\pm$0.7& 28.3\scriptsize$\pm$0.5& 33.1\scriptsize$\pm$1.8& 98.4\scriptsize$\pm$0.6\\
        \salun{} & \checkmark & 96.4\scriptsize$\pm$0.1& 38.1\scriptsize$\pm$6.7& 95.8\scriptsize$\pm$0.2& 1.9\scriptsize$\pm$1.2& 25.9\scriptsize$\pm$1.1& 36.9\scriptsize$\pm$7.2& 97.1\scriptsize$\pm$0.4\\
        \scrub{} & \checkmark & 96.7\scriptsize$\pm$0.1& 41.9\scriptsize$\pm$1.5& 95.9\scriptsize$\pm$0.1& 1.7\scriptsize$\pm$0.1& 24.9\scriptsize$\pm$0.2& 39.3\scriptsize$\pm$0.9& 97.4\scriptsize$\pm$0.2\\
        \rowmethod
        \method{} & \checkmark & 96.7\scriptsize$\pm$0.1& 35.2\scriptsize$\pm$2.0& 95.8\scriptsize$\pm$0.0& 4.2\scriptsize$\pm$1.1& 26.9\scriptsize$\pm$1.4& 34.3\scriptsize$\pm$2.7& 98.8\scriptsize$\pm$0.4\\
    \end{tabularx}
    \label{tab:celeba_0.9}
\end{table*}

\begin{table*}[t]
    \scriptsize
    \centering
    \caption{\textbf{Group-robust machine unlearning in \waterbird{} with 0.1 unlearning ratio.}
    We build the forget set by sampling data points from a single group. The unlearning ratio is set to 0.1. We compare \inlinecolorbox{colormethod}{\method{}} against \sparse{}, \salun{}, and \scrub{}. The \gap{} is computed against \retrain{} + \reweight{}.}
    \vspace{\baselineskip}
    \begin{tabularx}{\textwidth} {
        L{4cm}
        C{0.5cm}
        *{6}{Y}
        C{1.5cm}
    }
        method & RW & RA & UA & TA & MIA & EO & GA & \textbf{\gap{}} $\uparrow$ \\
    \toprule
        \pretrain{} & $\times$ & 99.0\scriptsize$\pm$0.1 & 73.3\scriptsize$\pm$18.9 & 88.0\scriptsize$\pm$0.5 & 53.3\scriptsize$\pm$9.4 & 25.9\scriptsize$\pm$0.7 & 57.6\scriptsize$\pm$2.0 & - \\
        \retrain{} & $\times$ & 99.0\scriptsize$\pm$0.2 & 46.7\scriptsize$\pm$24.9 & 87.1\scriptsize$\pm$0.8 & 60.0\scriptsize$\pm$28.3 & 27.4\scriptsize$\pm$1.7 & 57.4\scriptsize$\pm$2.9 & - \\
        \retrain{} & \checkmark{} & 98.7\scriptsize$\pm$0.3 & 46.7\scriptsize$\pm$24.9 & 87.3\scriptsize$\pm$0.7 & 66.7\scriptsize$\pm$24.9 & 26.7\scriptsize$\pm$1.3 & 57.6\scriptsize$\pm$4.8 & - \\
    \cmidrule(lr){1-9}
        \sparse{} & $\times$ & 99.0\scriptsize$\pm$0.1 & 60.0\scriptsize$\pm$16.3 & 85.3\scriptsize$\pm$0.8 & 46.7\scriptsize$\pm$24.9 & 29.2\scriptsize$\pm$1.7 & 57.9\scriptsize$\pm$2.6 & 92.3\scriptsize$\pm$3.0 \\
        \salun{} & $\times$ & 100.0\scriptsize$\pm$0.0 & 53.3\scriptsize$\pm$9.4 & 78.3\scriptsize$\pm$3.3 & 66.7\scriptsize$\pm$9.4 & 42.7\scriptsize$\pm$5.1 & 33.3\scriptsize$\pm$7.3 & 81.6\scriptsize$\pm$6.7 \\
        \scrub{} & $\times$ & 98.9\scriptsize$\pm$0.1 & 53.3\scriptsize$\pm$9.4 & 86.9\scriptsize$\pm$0.4 & 53.3\scriptsize$\pm$18.9 & 31.1\scriptsize$\pm$1.1 & 44.3\scriptsize$\pm$3.5 & 91.3\scriptsize$\pm$2.0 \\
        \rowmethod
        \method{} & $\times$ & 99.1\scriptsize$\pm$0.0 & 60.0\scriptsize$\pm$16.3 & 87.1\scriptsize$\pm$0.2 & 33.3\scriptsize$\pm$18.9 & 26.1\scriptsize$\pm$2.1 & 59.2\scriptsize$\pm$5.3 & 91.2\scriptsize$\pm$9.1 \\
    \cmidrule(lr){1-9}
        \sparse{} & \checkmark{} & 99.0\scriptsize$\pm$0.1 & 73.3\scriptsize$\pm$18.9 & 85.3\scriptsize$\pm$1.2 & 53.3\scriptsize$\pm$24.9 & 29.1\scriptsize$\pm$2.8 & 58.2\scriptsize$\pm$3.8 & 92.2\scriptsize$\pm$3.0 \\
        \salun{} & \checkmark{} & 99.9\scriptsize$\pm$0.1 & 53.3\scriptsize$\pm$9.4 & 76.7\scriptsize$\pm$2.9 & 86.7\scriptsize$\pm$9.4 & 45.9\scriptsize$\pm$4.7 & 29.6\scriptsize$\pm$7.4 & 81.3\scriptsize$\pm$2.9 \\
        \scrub{} & \checkmark{} & 98.8\scriptsize$\pm$0.2 & 53.3\scriptsize$\pm$9.4 & 86.8\scriptsize$\pm$0.7 & 60.0\scriptsize$\pm$16.3 & 31.5\scriptsize$\pm$0.8 & 42.8\scriptsize$\pm$3.5 & 89.8\scriptsize$\pm$0.4 \\
        \rowmethod
        \method{} & \checkmark{} & 100.0\scriptsize$\pm$0.0 & 73.3\scriptsize$\pm$18.9 & 87.3\scriptsize$\pm$0.3 & 73.3\scriptsize$\pm$24.9 & 26.2\scriptsize$\pm$0.3 & 61.7\scriptsize$\pm$0.8 & 93.3\scriptsize$\pm$0.7 \\
    \end{tabularx}
    \label{tab:waterbirds_0.1}
\end{table*}

\begin{table*}[t]
    \scriptsize
    \centering
    \caption{\textbf{Group-robust machine unlearning in \waterbird{} with 0.5 unlearning ratio.} We build the forget set by sampling data points from a single group. The unlearning ratio is set to 0.5. We compare \inlinecolorbox{colormethod}{\method{}} against \sparse{}, \salun{}, and \scrub{}. The \gap{} is computed against \retrain{} + \reweight{}.}
    \vspace{\baselineskip}
    \begin{tabularx}{\textwidth} {
        L{4cm}
        C{0.5cm}
        *{6}{Y}
        C{1.5cm}
    }
        method & RW & RA & UA & TA & MIA & EO & GA & \textbf{\gap{}} $\uparrow$ \\
    \toprule
        {\pretrain{}} & {$\times$} & {98.9\scriptsize$\pm$0.3} & {84.5\scriptsize$\pm$1.7} & {87.7\scriptsize$\pm$0.5} & {33.3\scriptsize$\pm$6.1} & {26.2\scriptsize$\pm$1.9} & {56.6\scriptsize$\pm$6.0} & {-} \\
        {\retrain{}} & {$\times$} & {98.7\scriptsize$\pm$0.3} & {52.4\scriptsize$\pm$8.9} & {86.5\scriptsize$\pm$0.2} & {54.8\scriptsize$\pm$9.4} & {30.4\scriptsize$\pm$0.5} & {49.4\scriptsize$\pm$1.6} & {-} \\
        \retrain{} & \checkmark & 99.0\scriptsize$\pm$0.1 & 59.5\scriptsize$\pm$11.8 & 87.2\scriptsize$\pm$0.3 & 53.6\scriptsize$\pm$8.7 & 28.3\scriptsize$\pm$2.0 & 51.6\scriptsize$\pm$6.0 & - \\
    \cmidrule(lr){1-9}
        \sparse{} & $\times$ & 99.0\scriptsize$\pm$0.1 & 59.5\scriptsize$\pm$8.9 & 85.6\scriptsize$\pm$0.4 & 44.0\scriptsize$\pm$11.8 & 32.2\scriptsize$\pm$1.8 & 48.8\scriptsize$\pm$7.4 & 94.4\scriptsize$\pm$0.3 \\
        \salun{} & $\times$ & 100.0\scriptsize$\pm$0.0 & 50.0\scriptsize$\pm$5.1 & 81.8\scriptsize$\pm$0.4 & 90.5\scriptsize$\pm$3.4 & 38.7\scriptsize$\pm$1.2 & 39.3\scriptsize$\pm$3.3 & 87.4\scriptsize$\pm$3.5 \\
        \scrub{} & $\times$ & 98.8\scriptsize$\pm$0.2 & 60.7\scriptsize$\pm$7.7 & 86.9\scriptsize$\pm$0.6 & 45.2\scriptsize$\pm$8.9 & 31.9\scriptsize$\pm$1.8 & 41.7\scriptsize$\pm$1.7 & 93.9\scriptsize$\pm$1.4 \\
        \rowmethod
        \method{} & $\times$ & 100.0\scriptsize$\pm$0.0 & 53.6\scriptsize$\pm$7.7 & 86.1\scriptsize$\pm$1.0 & 58.3\scriptsize$\pm$8.9 & 28.3\scriptsize$\pm$1.7 & 53.8\scriptsize$\pm$2.6 & 95.3\scriptsize$\pm$0.7 \\
    \cmidrule(lr){1-9}
        \sparse{} & \checkmark & 98.7\scriptsize$\pm$0.1 & 64.3\scriptsize$\pm$5.8 & 85.0\scriptsize$\pm$1.2 & 46.4\scriptsize$\pm$12.7 & 30.6\scriptsize$\pm$1.4 & 53.7\scriptsize$\pm$4.3 & 94.7\scriptsize$\pm$1.1 \\
        \salun{} & \checkmark & 100.0\scriptsize$\pm$0.0 & 47.6\scriptsize$\pm$4.5 & 81.1\scriptsize$\pm$1.9 & 91.7\scriptsize$\pm$7.3 & 39.0\scriptsize$\pm$2.2 & 39.0\scriptsize$\pm$1.6 & 85.8\scriptsize$\pm$4.2 \\
        \scrub{} & \checkmark & 98.9\scriptsize$\pm$0.2 & 66.7\scriptsize$\pm$1.7 & 87.0\scriptsize$\pm$0.5 & 44.0\scriptsize$\pm$8.9 & 30.9\scriptsize$\pm$1.3 & 44.3\scriptsize$\pm$3.1 & 94.5\scriptsize$\pm$1.3 \\
        \rowmethod
        \method{} & \checkmark & 99.9\scriptsize$\pm$0.1 & 54.8\scriptsize$\pm$14.7 & 85.8\scriptsize$\pm$0.7 & 59.5\scriptsize$\pm$12.1 & 28.3\scriptsize$\pm$2.9 & 53.7\scriptsize$\pm$3.8 & 96.9\scriptsize$\pm$1.6 \\
    \end{tabularx}
    \label{tab:waterbirds_0.5}
\end{table*}

\begin{table*}[ht]
    \scriptsize
    \centering
    \caption{\textbf{Group-robust machine unlearning in \waterbird{} with 0.9 unlearning ratio.}
    We build the forget set by sampling data points from a single group. The unlearning ratio is set to 0.9. We compare \inlinecolorbox{colormethod}{\method{}} against \sparse{}, \salun{}, and \scrub{}. The \gap{} is computed against \retrain{} + \reweight{}.}
    \vspace{\baselineskip}
    \begin{tabularx}{\textwidth} {
        L{4cm}
        C{0.5cm}
        *{6}{Y}
        C{1.5cm}
    }
        method & RW & RA & UA & TA & MIA & EO & GA & \textbf{\gap{}} $\uparrow$ \\
    \toprule
        \pretrain{} & $\times$ & 98.6\scriptsize$\pm$0.6 & 76.0\scriptsize$\pm$9.1 & 86.5\scriptsize$\pm$0.4 & 44.7\scriptsize$\pm$4.1 & 28.3\scriptsize$\pm$1.9 & 55.9\scriptsize$\pm$5.2 & - \\
        \retrain{} & $\times$ & 98.9\scriptsize$\pm$0.2 & 41.3\scriptsize$\pm$5.7 & 84.3\scriptsize$\pm$0.3 & 68.7\scriptsize$\pm$6.8 & 36.4\scriptsize$\pm$1.4 & 41.7\scriptsize$\pm$3.9 & - \\
        \retrain{} & \checkmark{} & 98.9\scriptsize$\pm$0.1 & 41.3\scriptsize$\pm$2.5 & 85.7\scriptsize$\pm$0.2 & 62.7\scriptsize$\pm$3.4 & 33.5\scriptsize$\pm$1.3 & 43.0\scriptsize$\pm$2.9 & - \\
    \cmidrule(lr){1-9}
        \sparse{} & $\times$ & 98.9\scriptsize$\pm$0.2 & 60.0\scriptsize$\pm$3.3 & 82.9\scriptsize$\pm$1.2 & 50.7\scriptsize$\pm$5.0 & 35.3\scriptsize$\pm$0.4 & 49.9\scriptsize$\pm$4.6 & 92.9\scriptsize$\pm$1.0 \\
        \salun{} & $\times$ & 100.0\scriptsize$\pm$0.0 & 40.0\scriptsize$\pm$4.3 & 81.3\scriptsize$\pm$0.9 & 92.7\scriptsize$\pm$3.4 & 41.4\scriptsize$\pm$1.3 & 30.8\scriptsize$\pm$2.2 & 89.6\scriptsize$\pm$2.0 \\
        \scrub{} & $\times$ & 97.8\scriptsize$\pm$0.1 & 30.7\scriptsize$\pm$1.9 & 86.1\scriptsize$\pm$0.5 & 52.7\scriptsize$\pm$3.4 & 36.6\scriptsize$\pm$1.0 & 25.1\scriptsize$\pm$1.5 & 92.7\scriptsize$\pm$0.7 \\
        \rowmethod
        \method{} & $\times$ & 100.0\scriptsize$\pm$0.0 & 66.7\scriptsize$\pm$5.7 & 85.7\scriptsize$\pm$0.7 & 58.7\scriptsize$\pm$4.7 & 32.1\scriptsize$\pm$1.5 & 49.8\scriptsize$\pm$3.3 & 93.4\scriptsize$\pm$1.3 \\
    \cmidrule(lr){1-9}
        \sparse{} & \checkmark{} & 99.0\scriptsize$\pm$0.2 & 59.3\scriptsize$\pm$10.6 & 84.6\scriptsize$\pm$0.6 & 45.3\scriptsize$\pm$6.8 & 31.5\scriptsize$\pm$0.7 & 55.0\scriptsize$\pm$4.0 & 91.5\scriptsize$\pm$4.1 \\
        \salun{} & \checkmark{} & 100.0\scriptsize$\pm$0.0 & 45.3\scriptsize$\pm$2.5 & 80.3\scriptsize$\pm$0.7 & 87.3\scriptsize$\pm$1.9 & 41.8\scriptsize$\pm$0.5 & 31.9\scriptsize$\pm$4.8 & 90.9\scriptsize$\pm$1.0 \\
        \scrub{} & \checkmark{} & 98.0\scriptsize$\pm$0.1 & 33.3\scriptsize$\pm$3.4 & 86.2\scriptsize$\pm$0.7 & 54.7\scriptsize$\pm$5.2 & 35.9\scriptsize$\pm$1.5 & 28.0\scriptsize$\pm$3.4 & 93.6\scriptsize$\pm$1.2 \\
        \rowmethod
        \method{} & \checkmark{} & 98.9\scriptsize$\pm$0.2 & 44.7\scriptsize$\pm$3.4 & 83.1\scriptsize$\pm$1.3 & 65.3\scriptsize$\pm$3.4 & 35.7\scriptsize$\pm$2.2 & 45.0\scriptsize$\pm$1.7 & 97.2\scriptsize$\pm$0.3 \\
    \end{tabularx}
    \label{tab:waterbirds_0.9}
\end{table*}

\begin{table*}[tp]
    \scriptsize
    \centering
    \caption{\textbf{Group-robust machine unlearning in \fairface{} with 0.1 unlearning ratio.}
    We build the forget set by sampling data points from a single group. The unlearning ratio is set to 0.1. We compare \inlinecolorbox{colormethod}{\method{}} against \sparse{}, \salun{}, and \scrub{}. The \gap{} is computed against \retrain{} + \reweight{}.}
    \vspace{\baselineskip}
    \begin{tabularx}{\textwidth} {
        L{4cm}
        C{0.5cm}
        *{6}{Y}
        C{1.5cm}
    }
        method & RW & RA & UA & TA & MIA & EO & GA & \textbf{\gap{}} $\uparrow$ \\
    \toprule
        \pretrain{} & $\times$ & 66.2\scriptsize$\pm$0.7 & 79.2\scriptsize$\pm$2.5 & 57.2\scriptsize$\pm$0.1 & 0.7\scriptsize$\pm$0.2 & 5.8\scriptsize$\pm$0.2 & 71.6\scriptsize$\pm$2.1 & - \\
        \retrain{} & $\times$ & 67.3\scriptsize$\pm$0.1 & 71.7\scriptsize$\pm$0.8 & 57.0\scriptsize$\pm$0.4 & 1.0\scriptsize$\pm$0.8 & 5.4\scriptsize$\pm$1.1 & 69.0\scriptsize$\pm$2.8 & - \\
        \retrain{} & \checkmark{} & 66.8\scriptsize$\pm$0.1 & 72.0\scriptsize$\pm$1.7 & 56.8\scriptsize$\pm$0.4 & 0.9\scriptsize$\pm$0.5 & 4.3\scriptsize$\pm$0.6 & 71.1\scriptsize$\pm$0.6 & - \\
    \cmidrule(lr){1-9}
        \sparse{} & $\times$ & 63.7\scriptsize$\pm$0.3 & 78.9\scriptsize$\pm$3.5 & 56.1\scriptsize$\pm$0.8 & 0.0\scriptsize$\pm$0.0 & 5.5\scriptsize$\pm$2.6 & 69.7\scriptsize$\pm$2.2 & 97.3\scriptsize$\pm$0.7 \\
        \salun{} & $\times$ & 65.9\scriptsize$\pm$0.8 & 73.9\scriptsize$\pm$3.9 & 55.1\scriptsize$\pm$1.1 & 0.5\scriptsize$\pm$0.0 & 2.9\scriptsize$\pm$1.1 & 69.8\scriptsize$\pm$7.0 & 97.8\scriptsize$\pm$0.9 \\
        \scrub{} & $\times$ & 68.4\scriptsize$\pm$0.5 & 78.7\scriptsize$\pm$0.8 & 57.5\scriptsize$\pm$0.3 & 0.2\scriptsize$\pm$0.2 & 5.7\scriptsize$\pm$0.2 & 70.4\scriptsize$\pm$1.5 & 97.9\scriptsize$\pm$0.5 \\
        \rowmethod
        \method{} & $\times$ & 66.9\scriptsize$\pm$0.5 & 81.3\scriptsize$\pm$0.2 & 57.3\scriptsize$\pm$0.2 & 0.2\scriptsize$\pm$0.2 & 5.3\scriptsize$\pm$0.6 & 70.4\scriptsize$\pm$0.5 & 97.8\scriptsize$\pm$0.5 \\
    \cmidrule(lr){1-9}
        \sparse{} & \checkmark{} & 64.0\scriptsize$\pm$0.3 & 72.7\scriptsize$\pm$0.7 & 56.4\scriptsize$\pm$0.6 & 0.2\scriptsize$\pm$0.2 & 5.3\scriptsize$\pm$1.2 & 69.1\scriptsize$\pm$0.7 & 98.6\scriptsize$\pm$0.3 \\
        \salun{} & \checkmark{} & 66.2\scriptsize$\pm$0.4 & 80.1\scriptsize$\pm$1.5 & 55.3\scriptsize$\pm$0.4 & 0.2\scriptsize$\pm$0.2 & 4.7\scriptsize$\pm$0.9 & 73.3\scriptsize$\pm$4.6 & 97.1\scriptsize$\pm$0.3 \\
        \scrub{} & \checkmark{} & 68.4\scriptsize$\pm$0.5 & 79.2\scriptsize$\pm$1.0 & 57.5\scriptsize$\pm$0.4 & 0.2\scriptsize$\pm$0.2 & 5.6\scriptsize$\pm$1.1 & 70.9\scriptsize$\pm$1.7 & 97.8\scriptsize$\pm$0.6 \\
        \rowmethod
        \method{} & \checkmark{} & 67.4\scriptsize$\pm$0.5 & 82.3\scriptsize$\pm$1.3 & 57.6\scriptsize$\pm$0.3 & 0.0\scriptsize$\pm$0.0 & 6.0\scriptsize$\pm$0.7 & 71.2\scriptsize$\pm$0.5 & 97.6\scriptsize$\pm$0.8 \\
    \end{tabularx}
    \label{tab:fairface_0.1}
\end{table*}

\begin{table*}[tp]
    \scriptsize
    \centering
    \caption{\textbf{Group-robust machine unlearning in \fairface{} with 0.5 unlearning ratio.} We build the forget set by sampling data points from a single group. The unlearning ratio is set to 0.5. We compare \inlinecolorbox{colormethod}{\method{}} against \sparse{}, \salun{}, and \scrub{}. The \gap{} is computed against \retrain{} + \reweight{}.}
    \vspace{\baselineskip}
    \begin{tabularx}{\textwidth} {
        L{4cm}
        C{0.5cm}
        *{6}{Y}
        C{1.5cm}
    }
        method & RW & RA & UA & TA & MIA & EO & GA & \textbf{\gap{}} $\uparrow$ \\
    \toprule
        {\pretrain{}} & {$\times$} & {65.6\scriptsize$\pm$0.7} & {79.0\scriptsize$\pm$1.2} & {57.2\scriptsize$\pm$0.4} & {0.2\scriptsize$\pm$0.1} & {5.4\scriptsize$\pm$1.7} & {71.2\scriptsize$\pm$2.4} & {-} \\
        {\retrain{}} & {$\times$} & {66.8\scriptsize$\pm$0.4} & {57.8\scriptsize$\pm$3.3} & {56.5\scriptsize$\pm$0.1} & {0.9\scriptsize$\pm$0.2} & {9.2\scriptsize$\pm$0.9} & {58.7\scriptsize$\pm$3.0} & {-} \\
        \retrain{} & \checkmark & 66.7\scriptsize$\pm$0.2 & 69.3\scriptsize$\pm$0.5 & 56.7\scriptsize$\pm$0.2 & 0.7\scriptsize$\pm$0.5 & 5.6\scriptsize$\pm$1.5 & 69.6\scriptsize$\pm$0.7 & - \\
    \cmidrule(lr){1-9}
        \sparse{} & $\times$ & 64.0\scriptsize$\pm$0.3 & 74.1\scriptsize$\pm$1.2 & 56.9\scriptsize$\pm$0.5 & 0.2\scriptsize$\pm$0.1 & 6.1\scriptsize$\pm$0.7 & 69.4\scriptsize$\pm$0.7 & 98.3\scriptsize$\pm$0.2 \\
        \salun{} & $\times$ & 66.3\scriptsize$\pm$0.4 & 66.6\scriptsize$\pm$3.4 & 55.9\scriptsize$\pm$0.6 & 0.3\scriptsize$\pm$0.1 & 9.0\scriptsize$\pm$0.5 & 60.3\scriptsize$\pm$2.4 & 97.1\scriptsize$\pm$1.1 \\
        \scrub{} & $\times$ & 66.9\scriptsize$\pm$0.1 & 65.4\scriptsize$\pm$1.6 & 56.7\scriptsize$\pm$0.7 & 1.0\scriptsize$\pm$0.0 & 9.9\scriptsize$\pm$1.3 & 61.3\scriptsize$\pm$2.5 & 97.0\scriptsize$\pm$0.5 \\
        \rowmethod
        \method{} & $\times$ & 66.7\scriptsize$\pm$0.2 & 74.7\scriptsize$\pm$1.2 & 57.2\scriptsize$\pm$0.7 & 0.3\scriptsize$\pm$0.0 & 6.0\scriptsize$\pm$2.0 & 66.1\scriptsize$\pm$4.4 & 98.1\scriptsize$\pm$0.4 \\
    \cmidrule(lr){1-9}
        \sparse{} & \checkmark & 64.4\scriptsize$\pm$0.1 & 72.9\scriptsize$\pm$2.1 & 56.0\scriptsize$\pm$0.9 & 0.3\scriptsize$\pm$0.1 & 6.1\scriptsize$\pm$2.1 & 67.0\scriptsize$\pm$6.8 & 97.3\scriptsize$\pm$0.3 \\
        \salun{} & \checkmark & 65.1\scriptsize$\pm$0.4 & 69.8\scriptsize$\pm$6.3 & 54.8\scriptsize$\pm$0.6 & 0.3\scriptsize$\pm$0.2 & 6.6\scriptsize$\pm$2.1 & 63.7\scriptsize$\pm$3.4 & 97.2\scriptsize$\pm$0.4 \\
        \scrub{} & \checkmark & 66.7\scriptsize$\pm$0.1 & 73.4\scriptsize$\pm$2.2 & 57.2\scriptsize$\pm$0.5 & 0.7\scriptsize$\pm$0.3 & 6.2\scriptsize$\pm$1.1 & 70.2\scriptsize$\pm$2.7 & 98.7\scriptsize$\pm$0.7 \\
        \rowmethod
        \method{} & \checkmark & 64.7\scriptsize$\pm$0.3 & 71.6\scriptsize$\pm$2.8 & 57.1\scriptsize$\pm$0.3 & 0.3\scriptsize$\pm$0.2 & 5.8\scriptsize$\pm$0.4 & 70.3\scriptsize$\pm$1.6 & 98.7\scriptsize$\pm$0.8 \\
    \end{tabularx}
    \label{tab:fairface_0.5}
\end{table*}

\begin{table*}[tp]
    \scriptsize
    \centering
    \caption{\textbf{Group-robust machine unlearning in \fairface{} with 0.9 unlearning ratio.}
    We build the forget set by sampling data points from a single group. The unlearning ratio is set to 0.9. We compare \inlinecolorbox{colormethod}{\method{}} against \sparse{}, \salun{}, and \scrub{}. The \gap{} is computed against \retrain{} + \reweight{}.}
    \vspace{\baselineskip}
    \begin{tabularx}{\textwidth} {
        L{4cm}
        C{0.5cm}
        *{6}{Y}
        C{1.5cm}
    }
        method & RW & RA & UA & TA & MIA & EO & GA & \textbf{\gap{}} $\uparrow$ \\
    \toprule
        \pretrain{} & $\times$ & 66.0\scriptsize$\pm$0.0 & 77.5\scriptsize$\pm$2.1 & 56.6\scriptsize$\pm$0.4 & 0.2\scriptsize$\pm$0.0 & 5.5\scriptsize$\pm$1.1 & 69.0\scriptsize$\pm$2.8 & - \\
        \retrain{} & $\times$ & 67.3\scriptsize$\pm$0.6 & 38.5\scriptsize$\pm$1.8 & 56.0\scriptsize$\pm$0.4 & 2.7\scriptsize$\pm$0.3 & 23.1\scriptsize$\pm$1.5 & 37.1\scriptsize$\pm$2.1 & - \\
        \retrain{} & \checkmark{} & 67.1\scriptsize$\pm$0.4 & 53.6\scriptsize$\pm$1.2 & 56.6\scriptsize$\pm$0.4 & 1.8\scriptsize$\pm$0.1 & 11.9\scriptsize$\pm$1.0 & 53.9\scriptsize$\pm$0.7 & - \\
    \cmidrule(lr){1-9}
        \sparse{} & $\times$ & 64.5\scriptsize$\pm$0.2 & 57.1\scriptsize$\pm$2.1 & 55.2\scriptsize$\pm$0.6 & 0.4\scriptsize$\pm$0.1 & 13.0\scriptsize$\pm$0.7 & 51.0\scriptsize$\pm$1.9 & 97.7\scriptsize$\pm$0.3 \\
        \salun{} & $\times$ & 65.7\scriptsize$\pm$0.5 & 46.5\scriptsize$\pm$6.2 & 53.9\scriptsize$\pm$0.1 & 0.5\scriptsize$\pm$0.1 & 15.3\scriptsize$\pm$1.0 & 42.8\scriptsize$\pm$4.9 & 95.2\scriptsize$\pm$1.8 \\
        \scrub{} & $\times$ & 60.2\scriptsize$\pm$1.0 & 52.7\scriptsize$\pm$4.4 & 53.3\scriptsize$\pm$0.5 & 2.4\scriptsize$\pm$0.7 & 15.6\scriptsize$\pm$1.4 & 48.7\scriptsize$\pm$4.9 & 95.7\scriptsize$\pm$0.6 \\
        \rowmethod
        \method{} & $\times$ & 68.2\scriptsize$\pm$0.3 & 64.4\scriptsize$\pm$2.5 & 56.5\scriptsize$\pm$0.5 & 0.5\scriptsize$\pm$0.1 & 10.4\scriptsize$\pm$0.7 & 56.8\scriptsize$\pm$2.6 & 97.0\scriptsize$\pm$1.0 \\
    \cmidrule(lr){1-9}
        \sparse{} & \checkmark{} & 64.0\scriptsize$\pm$0.4 & 74.5\scriptsize$\pm$3.0 & 55.9\scriptsize$\pm$0.4 & 0.3\scriptsize$\pm$0.2 & 5.6\scriptsize$\pm$0.6 & 69.8\scriptsize$\pm$4.6 & 91.9\scriptsize$\pm$1.5 \\
        \salun{} & \checkmark{} & 65.5\scriptsize$\pm$0.5 & 66.1\scriptsize$\pm$3.7 & 55.3\scriptsize$\pm$0.2 & 0.5\scriptsize$\pm$0.3 & 7.2\scriptsize$\pm$1.2 & 62.8\scriptsize$\pm$4.9 & 94.9\scriptsize$\pm$1.7 \\
        \scrub{} & \checkmark{} & 61.2\scriptsize$\pm$1.1 & 65.5\scriptsize$\pm$3.5 & 54.5\scriptsize$\pm$0.3 & 1.3\scriptsize$\pm$0.1 & 9.5\scriptsize$\pm$0.9 & 64.4\scriptsize$\pm$2.7 & 94.5\scriptsize$\pm$1.6 \\
        \rowmethod
        \method{} & \checkmark{} & 64.7\scriptsize$\pm$0.2 & 67.1\scriptsize$\pm$1.4 & 56.7\scriptsize$\pm$0.2 & 0.5\scriptsize$\pm$0.1 & 8.8\scriptsize$\pm$0.2 & 63.5\scriptsize$\pm$1.6 & 94.9\scriptsize$\pm$0.6 \\
    \end{tabularx}
    \label{tab:fairface_0.9}
\end{table*}

\subsection{Impact of different unlearning ratios}
\label{appxsub:ratios}
For completeness purposes, this section reports all tables associated with \cref{fig:ratio,fig:ratio_full,fig:robust_ratio,fig:reweight}.
For these experiments, the main paper summarizes the results to reduce the occupied space and simplify the interpretation (\ie, limiting the reported metrics to one).
Therefore, \cref{tab:celeba_0.1,tab:celeba_0.9,tab:waterbirds_0.1,tab:waterbirds_0.9,tab:fairface_0.1,tab:fairface_0.9} present experiments in \celeba{}, \waterbird{}, and \fairface{} datasets, with unlearning ratios of $0.1$, and $0.9$.
Additionally, \cref{tab:celeba_0.5,tab:waterbirds_0.5,tab:fairface_0.5} report the standard deviations of \cref{tab:robust_celeba,tab:robust_waterbirds,tab:robust_fairface}.

Overall, we notice that in \celeba{}, the higher the unlearning ratio, the lower the forget accuracy (from 43.0\% to 32.6\% using \method{}), with the gap being reduced when \reweight{} is included in the retaining step (from 44.9\% to 35.2\% with \method{}).
RA, TA, and MIA remain stable across different unlearning ratios.
Instead, EO behaves similarly to GA and UA, with high values at high unlearning ratios (\eg, \method{} scores 22.9\% \vs{} 26.9\% with 0.1 and 0.9 unlearning ratios).

In \waterbird{}, metrics do not show global trends, except for EO and GA, which worsen as the unlearning ratio grows.
Therefore, methods lose accuracy on the dominant group of the forget set as the ratio of unlearned samples grows, with \scrub{} showing the highest drop (from 44.3\% to 25.1\%).
Similarly, GA and EO worsen even when adding \reweight{}.
As the unlearning ratio grows, the number of forget set dominant group samples left in the remaining set lowers.
Given the small size of \waterbird{} (4795 samples, of which 56 are in the smallest group), \reweight{} strongly increases the sampling likelihood of a restricted number of samples to preserve original accuracy, causing overfitting.
Thus, the overall benefit of \reweight{} is reduced (GA increases by only 1.3 with \retrain{}).

Also \fairface{} shows a general drop in UA, which grows when \reweight{} is applied.
EO and GA also behave like in \celeba{}, with an enhanced degradation at higher unlearning ratios.
However, \method{} shows good robustness even without \reweight{}, scoring EO and GA that are close to \retrain{}+\reweight{}, \eg, 10.4\% \vs{} 11.9\% in EO and 56.8\% \vs{} 53.9 in GA with a 0.9 unlearning ratio (\cref{tab:fairface_0.9}). 

\begin{table*}[tp]
    \scriptsize
    \centering
    \caption{\textbf{Group-robust machine unlearning in \fairface{} by sampling from 9 groups.} We build the forget set by sampling data points from 9 groups. The unlearning ratio is set to 0.5. We compare \inlinecolorbox{colormethod}{\method{}} against \sparse{}, \salun{}, and \scrub{}. The \gap{} is computed against \retrain{} + \reweight{}.}
    \vspace{\baselineskip}
    \begin{tabularx}{\textwidth} {
        L{4cm}
        C{0.5cm}
        *{6}{Y}
        C{1.5cm}
    }
        method & RW & RA & UA & TA & MIA & EO & GA & \textbf{\gap{}} $\uparrow$ \\
    \toprule
        \pretrain{} & $\times$ & 64.6\scriptsize$\pm$0.5 & 81.5\scriptsize$\pm$0.5 & 57.4\scriptsize$\pm$0.3 & 0.4\scriptsize$\pm$0.0 & 1.1\scriptsize$\pm$0.1 & 72.5\scriptsize$\pm$0.7 & - \\
        \retrain{} & $\times$ & 66.5\scriptsize$\pm$0.5 & 60.1\scriptsize$\pm$0.9 & 55.4\scriptsize$\pm$0.4 & 1.1\scriptsize$\pm$0.1 & 5.6\scriptsize$\pm$0.4 & 60.4\scriptsize$\pm$1.7 & - \\
        \retrain{} & \checkmark{} & 64.4\scriptsize$\pm$1.1 & 72.1\scriptsize$\pm$2.2 & 56.3\scriptsize$\pm$0.3 & 0.5\scriptsize$\pm$0.0 & 2.0\scriptsize$\pm$0.6 & 71.8\scriptsize$\pm$2.7 & - \\
    \cmidrule(lr){1-9}
        \sparse{} & $\times$ & 63.5\scriptsize$\pm$0.6 & 69.9\scriptsize$\pm$3.9 & 55.3\scriptsize$\pm$0.1 & 0.5\scriptsize$\pm$0.0 & 2.6\scriptsize$\pm$1.0 & 64.4\scriptsize$\pm$4.4 & 97.7\scriptsize$\pm$1.1 \\
        \salun{} & $\times$ & 64.3\scriptsize$\pm$0.5 & 64.6\scriptsize$\pm$1.2 & 54.0\scriptsize$\pm$0.1 & 0.2\scriptsize$\pm$0.1 & 3.6\scriptsize$\pm$0.3 & 59.7\scriptsize$\pm$1.5 & 96.0\scriptsize$\pm$0.6 \\
        \scrub{} & $\times$ & 67.2\scriptsize$\pm$0.4 & 74.3\scriptsize$\pm$0.7 & 56.9\scriptsize$\pm$0.2 & 0.3\scriptsize$\pm$0.1 & 1.8\scriptsize$\pm$0.7 & 65.6\scriptsize$\pm$0.6 & 97.7\scriptsize$\pm$0.3 \\
        \rowmethod
        \method{} & $\times$ & 66.3\scriptsize$\pm$0.4 & 74.2\scriptsize$\pm$0.4 & 56.8\scriptsize$\pm$0.5 & 0.3\scriptsize$\pm$0.1 & 1.7\scriptsize$\pm$0.4 & 65.7\scriptsize$\pm$0.6 & 97.9\scriptsize$\pm$0.4 \\
    \cmidrule(lr){1-9}
        \sparse{} & \checkmark{} & 63.7\scriptsize$\pm$0.1 & 75.2\scriptsize$\pm$0.6 & 56.3\scriptsize$\pm$0.1 & 0.2\scriptsize$\pm$0.1 & 1.3\scriptsize$\pm$0.3 & 69.6\scriptsize$\pm$1.2 & 98.6\scriptsize$\pm$0.2 \\
        \salun{} & \checkmark{} & 63.7\scriptsize$\pm$0.8 & 74.4\scriptsize$\pm$1.5 & 55.5\scriptsize$\pm$0.4 & 0.4\scriptsize$\pm$0.0 & 2.3\scriptsize$\pm$0.3 & 69.0\scriptsize$\pm$1.6 & 98.4\scriptsize$\pm$0.2 \\
        \scrub{} & \checkmark{} & 66.7\scriptsize$\pm$0.4 & 80.5\scriptsize$\pm$0.1 & 57.4\scriptsize$\pm$0.5 & 0.4\scriptsize$\pm$0.1 & 1.7\scriptsize$\pm$0.3 & 71.3\scriptsize$\pm$0.4 & 97.4\scriptsize$\pm$0.3 \\
        \rowmethod
        \method{} & \checkmark{} & 63.4\scriptsize$\pm$0.4 & 73.2\scriptsize$\pm$0.3 & 56.7\scriptsize$\pm$0.3 & 0.4\scriptsize$\pm$0.1 & 1.5\scriptsize$\pm$0.5 & 70.3\scriptsize$\pm$1.0 & 99.0\scriptsize$\pm$0.1 \\
    \end{tabularx}
    \label{tab:fairface_3g}
\end{table*}

\begin{table*}[t]
    \scriptsize
    \centering
    \caption{\textbf{Group-robust machine unlearning in \fairface{} by sampling from 25 groups.} We build the forget set by sampling data points from 25 groups. The unlearning ratio is set to 0.5. We compare \inlinecolorbox{colormethod}{\method{}} against \sparse{}, \salun{}, and \scrub{}. The \gap{} is computed against \retrain{} + \reweight{}.}
    \vspace{\baselineskip}
    \begin{tabularx}{\textwidth} {
        L{4cm}
        C{0.5cm}
        *{6}{Y}
        C{1.5cm}
    }
        method & RW & RA & UA & TA & MIA & EO & GA & \textbf{\gap{}} $\uparrow$ \\
    \toprule
        \pretrain{} & $\times$ & 65.1\scriptsize$\pm$0.3 & 70.6\scriptsize$\pm$0.5 & 56.9\scriptsize$\pm$0.3 & 0.3\scriptsize$\pm$0.1 & 1.6\scriptsize$\pm$0.6 & 62.2\scriptsize$\pm$1.4 & - \\
        \retrain{} & $\times$ & 66.5\scriptsize$\pm$1.0 & 55.5\scriptsize$\pm$2.1 & 54.8\scriptsize$\pm$0.7 & 0.8\scriptsize$\pm$0.1 & 1.9\scriptsize$\pm$0.3 & 56.0\scriptsize$\pm$2.1 & - \\
        \retrain{} & \checkmark{} & 66.1\scriptsize$\pm$0.6 & 60.5\scriptsize$\pm$0.7 & 55.5\scriptsize$\pm$0.4 & 0.7\scriptsize$\pm$0.1 & 2.3\scriptsize$\pm$0.3 & 60.5\scriptsize$\pm$0.5 & - \\
    \cmidrule(lr){1-9}
        \sparse{} & $\times$ & 65.5\scriptsize$\pm$0.8 & 61.8\scriptsize$\pm$0.3 & 55.5\scriptsize$\pm$0.5 & 0.5\scriptsize$\pm$0.1 & 1.6\scriptsize$\pm$0.8 & 57.0\scriptsize$\pm$0.9 & 98.7\scriptsize$\pm$0.2 \\
        \salun{} & $\times$ & 64.5\scriptsize$\pm$0.5 & 60.0\scriptsize$\pm$3.8 & 55.2\scriptsize$\pm$1.0 & 0.5\scriptsize$\pm$0.1 & 1.1\scriptsize$\pm$0.4 & 57.0\scriptsize$\pm$3.8 & 98.0\scriptsize$\pm$0.7 \\
        \scrub{} & $\times$ & 66.7\scriptsize$\pm$0.6 & 62.4\scriptsize$\pm$1.0 & 55.4\scriptsize$\pm$0.3 & 0.3\scriptsize$\pm$0.1 & 1.7\scriptsize$\pm$0.9 & 55.2\scriptsize$\pm$0.6 & 98.4\scriptsize$\pm$0.2 \\
        \rowmethod
        \method{} & $\times$ & 66.8\scriptsize$\pm$0.2 & 64.8\scriptsize$\pm$0.5 & 56.4\scriptsize$\pm$0.5 & 0.3\scriptsize$\pm$0.1 & 1.7\scriptsize$\pm$0.7 & 57.5\scriptsize$\pm$1.1 & 98.3\scriptsize$\pm$0.2 \\
    \cmidrule(lr){1-9}
        \sparse{} & \checkmark{} & 64.3\scriptsize$\pm$0.7 & 66.7\scriptsize$\pm$0.4 & 55.8\scriptsize$\pm$0.4 & 0.4\scriptsize$\pm$0.1 & 2.1\scriptsize$\pm$0.8 & 60.8\scriptsize$\pm$1.0 & 98.3\scriptsize$\pm$0.3 \\
        \salun{} & \checkmark{} & 62.5\scriptsize$\pm$1.6 & 64.6\scriptsize$\pm$2.0 & 54.6\scriptsize$\pm$0.8 & 0.3\scriptsize$\pm$0.1 & 2.2\scriptsize$\pm$0.4 & 60.1\scriptsize$\pm$2.0 & 98.0\scriptsize$\pm$0.3 \\
        \scrub{} & \checkmark{} & 65.3\scriptsize$\pm$0.3 & 71.4\scriptsize$\pm$0.6 & 57.0\scriptsize$\pm$0.3 & 0.2\scriptsize$\pm$0.0 & 1.9\scriptsize$\pm$0.9 & 63.7\scriptsize$\pm$0.7 & 97.1\scriptsize$\pm$0.1 \\
        \rowmethod
        \method{} & \checkmark{} & 66.2\scriptsize$\pm$1.6 & 63.0\scriptsize$\pm$1.9 & 54.3\scriptsize$\pm$0.6 & 0.9\scriptsize$\pm$0.3 & 3.5\scriptsize$\pm$0.9 & 57.6\scriptsize$\pm$3.3 & 98.3\scriptsize$\pm$0.1 \\
    \end{tabularx}
    \label{tab:fairface_5g}
\end{table*}

\subsection{Multi-group unlearning}
\label{appxsub:groups}
This section further expands the experimental protocol of \cref{sub:multigroup}, highlighting the sampling composition and reporting all investigated metrics.
For the 9 groups experiment, we sampled the forget set data with either \emph{\targetattr{20-29 y.o.}, \targetattr{50-59 y.o.}}, and \emph{\targetattr{3-9 y.o.\ }}target attributes, and \emph{\sensitiveattr{Afro-American}, \sensitiveattr{Latino-Hispanic}}, and \emph{\sensitiveattr{Caucasian}} sensitive attributes.
Instead, for the 25 groups experiment, we additionally sample from the \emph{\targetattr{more than 70 y.o.\ }}and \emph{\targetattr{30-39 y.o.\ }}target attributes, and \emph{\sensitiveattr{East Asian}} and \emph{\sensitiveattr{Middle Eastern}} sensitive attributes.
The choice of groups is random, and the unlearning ratio is fixed to 0.5. 

Full results for unlearning multiple groups in group-robust unlearning are reported in \cref{tab:fairface_3g,tab:fairface_5g}.
We notice overall the same trend as in \cref{fig:multi_group}.
\reweight{} contribution gets less important as the number of groups from which the forget set is sampled decreases, highlighted by unchanged UA, EO, and GA.
The GA, when using \method{}, \eg, increases by nearly 5\% in the 9 groups experiment (\cref{tab:fairface_3g}), while it remains unchanged in the 25 groups one (\cref{tab:fairface_5g}).
Test accuracies are better preserved when unlearning 1 or 9 groups, while we notice a drop (about 1.5\%) in the 25 groups experiment.
We argue that this decline is caused by the larger forget set, which reduces the number of available samples in the remaining set.
Overall, \method{} approximates \retrain{} + \reweight{} better than baselines, consistently achieving the best \gap{}.

\subsection{Unlearning on MultiNLI}
\begin{table*}[t]
    \scriptsize
    \centering
    \caption{\changed{\textbf{Group-robust machine unlearning in \multinli{} with 0.5 unlearning ratio.} We build the forget set by sampling data points from a single group. The unlearning ratio is set to 0.5. We compare \inlinecolorbox{colormethod}{\method{}} against \sparse{}, \salun{}, and \scrub{}. The \gap{} is computed against \retrain{} + \reweight{}.}}
    \vspace{\baselineskip}
    \begin{tabularx}{\textwidth} {
        L{4cm}
        C{0.5cm}
        *{6}{Y}
        C{1.5cm}
    }
        method & RW & RA & UA & TA & MIA & EO & GA & \textbf{\gap{}} $\uparrow$ \\
    \toprule
        \pretrain{} & $\times$ & 94.7\scriptsize$\pm$2.5  & 81.8\scriptsize$\pm$8.4  & 82.3\scriptsize$\pm$0.1  & 31.4\scriptsize$\pm$0.9  & 14.4\scriptsize$\pm$1.8  & 59.3\scriptsize$\pm$2.7  & - \\
        \retrain{} & $\times$ & 95.9\scriptsize$\pm$1.1  & 53.0\scriptsize$\pm$3.7  & 82.3\scriptsize$\pm$0.1  & 47.6\scriptsize$\pm$3.9  & 19.3\scriptsize$\pm$0.6  & 50.8\scriptsize$\pm$3.1  & - \\
        \retrain{} & \checkmark & 95.3\scriptsize$\pm$1.1  & 59.8\scriptsize$\pm$0.7  & 81.7\scriptsize$\pm$0.0  & 43.3\scriptsize$\pm$2.3  & 16.2\scriptsize$\pm$0.3  & 58.8\scriptsize$\pm$2.0  & - \\
    \cmidrule(lr){1-9}
        \sparse{} & $\times$ & 84.8\scriptsize$\pm$2.0 & 46.6\scriptsize$\pm$4.8 & 79.0\scriptsize$\pm$0.7 & 34.1\scriptsize$\pm$2.3 & 23.3\scriptsize$\pm$1.4 & 42.0\scriptsize$\pm$2.1 & 90.1\scriptsize$\pm$1.8\\
        \salun{} & $\times$ & 76.4\scriptsize$\pm$30.3 & 36.5\scriptsize$\pm$25.9 & 63.8\scriptsize$\pm$21.4 & 36.1\scriptsize$\pm$25.6 & 20.2\scriptsize$\pm$0.8 & 35.1\scriptsize$\pm$24.9 & 80.8\scriptsize$\pm$19.0\\
        \scrub{} & $\times$ & 96.4\scriptsize$\pm$1.9 & 56.8\scriptsize$\pm$4.6 & 82.4\scriptsize$\pm$0.2 & 46.7\scriptsize$\pm$5.5 & 22.6\scriptsize$\pm$1.7 & 47.6\scriptsize$\pm$3.3 & 95.3\scriptsize$\pm$1.3\\
        \rowmethod
        \method{} & $\times$ & 99.2\scriptsize$\pm$0.1 & 60.3\scriptsize$\pm$1.6 & 79.2\scriptsize$\pm$0.2 & 54.9\scriptsize$\pm$2.8 & 18.8\scriptsize$\pm$0.7 & 55.6\scriptsize$\pm$2.0 & 95.8\scriptsize$\pm$0.3\\
    \cmidrule(lr){1-9}
        \sparse{} & \checkmark & 84.9\scriptsize$\pm$2.0 & 55.4\scriptsize$\pm$3.8 & 79.0\scriptsize$\pm$0.6 & 31.7\scriptsize$\pm$1.4 & 19.3\scriptsize$\pm$1.1 & 49.2\scriptsize$\pm$1.6 & 93.0\scriptsize$\pm$1.2\\
        \salun{} & \checkmark & 96.0\scriptsize$\pm$0.1 & 54.3\scriptsize$\pm$3.6 & 78.3\scriptsize$\pm$0.2 & 48.1\scriptsize$\pm$2.0 & 18.5\scriptsize$\pm$1.9 & 51.9\scriptsize$\pm$2.6 & 95.9\scriptsize$\pm$2.3\\
        \scrub{} & \checkmark & 96.5\scriptsize$\pm$1.9 & 82.1\scriptsize$\pm$7.8 & 82.4\scriptsize$\pm$0.2 & 32.2\scriptsize$\pm$8.9 & 14.0\scriptsize$\pm$3.8 & 62.4\scriptsize$\pm$6.3 & 92.5\scriptsize$\pm$2.8\\
        \rowmethod
        \method{} & \checkmark & 99.9\scriptsize$\pm$0.0 & 67.0\scriptsize$\pm$2.6 & 81.5\scriptsize$\pm$0.2 & 51.1\scriptsize$\pm$2.6 & 18.0\scriptsize$\pm$0.4 & 56.6\scriptsize$\pm$1.2 & 96.0\scriptsize$\pm$0.8\\
        \end{tabularx}
        \label{tab:multinli_0.5}
\end{table*}

\changed{\Cref{tab:multinli_0.5} shows the result for group-robust unlearning on \multinli{} using an unlearning ratio $r$ of 0.5.
\reweight{} allows a successful recovery of the original group accuracy (-0.5), while preserving the test accuracy (-0.6).
Compared to \pretrain{}, the EO are worse (14.4 \vs{} 16.2) but better compared to plain \retrain{} (16.2 \vs 19.3), thus, \reweight{} partly recovered also the equalized odds.
\method{} achieves the best performance preservation overall by scoring the highest GA alignment, both when using (96.0) and not using (95.8) \reweight{}.
Particularly, \method{} is less influenced by \reweight{}, and most of the alignment comes from its unique loss formulation.
\scrub{} achieves the second-best result without \reweight{} (95.3), but struggles when used (92.5), as forget accuracy unexpectedly grows to high values (82.1).
Instead, \salun{} scores the best \gap{} after \method{} (95.9), but shows high variance and low results without \reweight{}.
Overall, \method{} is the most consistent among the two configurations and generally shows the lowest variance across metrics.}

\begin{table*}[t]
    \scriptsize
    \centering
    \caption{\changed{\textbf{\method{} ablations.} We compute \method{} ablations on each of the three investigated datasets. From left to right, we report the investigated dataset, the \emph{retaining term}, the \emph{unlearning term}, the \emph{calibration term}, and \reweight{}. We measure performance using all metrics. The configuration that corresponds to \inlinecolorbox{colormethod}{\method{} + \reweight{}} is highlighted.}}
    \vspace{\baselineskip}
    \begin{tabularx}{\linewidth} { 
        L{1.3cm}
        *{4}{C{0.65cm}}
        *{6}{Y}
        C{1.5cm}
    }
            dataset & \cref{eq:retain} & \cref{eq:unlearn} & \cref{eq:regularize} & RW & RA & UA & TA & MIA & EO & GA & \textbf{\gap{}} $\uparrow$ \\
        \toprule
            \multirow{4}{*}{CelebA}
             & \checkmark & \checkmark & $\times$ & $\times$ & 96.4\scriptsize$\pm$0.1& 35.5\scriptsize$\pm$0.9& 95.9\scriptsize$\pm$0.0& 1.1\scriptsize$\pm$0.1& 26.7\scriptsize$\pm$0.9& 35.4\scriptsize$\pm$2.3& 97.5\scriptsize$\pm$0.5\\
             & \checkmark & \checkmark & \checkmark & $\times$ & 96.4\scriptsize$\pm$0.1& 36.3\scriptsize$\pm$0.9& 95.9\scriptsize$\pm$0.1& 1.0\scriptsize$\pm$0.3& 26.1\scriptsize$\pm$0.8& 36.3\scriptsize$\pm$2.0& 98.0\scriptsize$\pm$0.4\\
             & $\times$ & \checkmark & \checkmark & $\times$ & 95.8\scriptsize$\pm$0.5& 27.8\scriptsize$\pm$5.7& 95.3\scriptsize$\pm$0.4& 0.9\scriptsize$\pm$0.2& 25.5\scriptsize$\pm$0.8& 28.5\scriptsize$\pm$6.9& 95.2\scriptsize$\pm$2.5\\
             & \cellcolor{colormethod}\checkmark & \cellcolor{colormethod}\checkmark & \cellcolor{colormethod}\checkmark & \cellcolor{colormethod}\checkmark & \cellcolor{colormethod}96.3\scriptsize$\pm$0.2 & \cellcolor{colormethod}43.2\scriptsize$\pm$0.6 & \cellcolor{colormethod}96.0\scriptsize$\pm$0.0 & \cellcolor{colormethod}1.2\scriptsize$\pm$0.1 & \cellcolor{colormethod}24.0\scriptsize$\pm$0.5 & \cellcolor{colormethod}41.3\scriptsize$\pm$1.1 & \cellcolor{colormethod}99.0\scriptsize$\pm$0.2 \\
        \cmidrule{1-12}
            \multirow{4}{*}{Waterbirds}
             & \checkmark & \checkmark & $\times$ & $\times$ & 100.0\scriptsize$\pm$0.0 & 47.6\scriptsize$\pm$7.3 & 85.0\scriptsize$\pm$0.6 & 73.8\scriptsize$\pm$3.4 & 32.3\scriptsize$\pm$0.8 & 51.1\scriptsize$\pm$0.6 & 92.5\scriptsize$\pm$4.6 \\
             & \checkmark & \checkmark & \checkmark & $\times$ & 100.0\scriptsize$\pm$0.0 & 53.6\scriptsize$\pm$7.7 & 86.1\scriptsize$\pm$1.0 & 58.3\scriptsize$\pm$8.9 & 28.3\scriptsize$\pm$1.7 & 53.8\scriptsize$\pm$2.6 & 95.3\scriptsize$\pm$0.7 \\
             & $\times$ & \checkmark & \checkmark & $\times$ & 93.0\scriptsize$\pm$3.3 & 16.7\scriptsize$\pm$9.4 & 80.3\scriptsize$\pm$2.9 & 64.3\scriptsize$\pm$12.7 & 35.8\scriptsize$\pm$7.8 & 16.8\scriptsize$\pm$8.9 & 81.9\scriptsize$\pm$5.5 \\
             & \cellcolor{colormethod}\checkmark & \cellcolor{colormethod}\checkmark & \cellcolor{colormethod}\checkmark & \cellcolor{colormethod}\checkmark & \cellcolor{colormethod}99.9\scriptsize$\pm$0.1 & \cellcolor{colormethod}54.8\scriptsize$\pm$14.7 & \cellcolor{colormethod}85.8\scriptsize$\pm$0.7 & \cellcolor{colormethod}59.5\scriptsize$\pm$12.1 & \cellcolor{colormethod}28.3\scriptsize$\pm$2.9 & \cellcolor{colormethod}53.7\scriptsize$\pm$3.8 & \cellcolor{colormethod}96.9\scriptsize$\pm$1.6 \\
        \cmidrule{1-12}
            \multirow{4}{*}{FairFace}
             & \checkmark & \checkmark & $\times$ & $\times$ & 65.2\scriptsize$\pm$0.1 & 63.1\scriptsize$\pm$1.6 & 56.9\scriptsize$\pm$0.3 & 0.3\scriptsize$\pm$0.0 & 10.3\scriptsize$\pm$0.8 & 59.2\scriptsize$\pm$1.9 & 96.1\scriptsize$\pm$0.8 \\
             & \checkmark & \checkmark & \checkmark & $\times$ & 66.7\scriptsize$\pm$0.2 & 74.7\scriptsize$\pm$1.2 & 57.2\scriptsize$\pm$0.7 & 0.3\scriptsize$\pm$0.0 & 6.0\scriptsize$\pm$2.0 & 66.1\scriptsize$\pm$4.4 & 98.1\scriptsize$\pm$0.4 \\
             & $\times$ & \checkmark & \checkmark & $\times$ & 59.1\scriptsize$\pm$3.1 & 87.1\scriptsize$\pm$6.8 & 54.5\scriptsize$\pm$1.8 & 0.0\scriptsize$\pm$0.0 & 3.1\scriptsize$\pm$0.5 & 81.1\scriptsize$\pm$6.2 & 93.0\scriptsize$\pm$3.1 \\
             & \cellcolor{colormethod}\checkmark & \cellcolor{colormethod}\checkmark & \cellcolor{colormethod}\checkmark & \cellcolor{colormethod}\checkmark & \cellcolor{colormethod}64.7\scriptsize$\pm$0.3 & \cellcolor{colormethod}71.6\scriptsize$\pm$2.8 & \cellcolor{colormethod}57.1\scriptsize$\pm$0.3 & \cellcolor{colormethod}0.3\scriptsize$\pm$0.2 & \cellcolor{colormethod}5.8\scriptsize$\pm$0.4 & \cellcolor{colormethod}70.3\scriptsize$\pm$1.6 & \cellcolor{colormethod}98.7\scriptsize$\pm$0.8 \\
    \end{tabularx}
    \label{tab:ablations_full}
\end{table*}

\subsection{Extended ablation study}
\label{appxsub:ablations}
\Cref{sub:ablations} shows a comprehensive ablation of \method{}'s components.
However, \cref{tab:ablations} limits the analysis only to the UA, GA, and \gap{} to reduce space usage.
Thus, \cref{tab:ablations_full} reports all metrics investigated for completeness.
Although \cref{tab:ablations} metrics are limited, we chose a subset that shows great variance along different components and is more interesting to evaluate.
For instance, EO variations are more nuanced compared to GA (\eg, it drops by 4.5 in \fairface{}, while GA grows by 11.1).
\begin{wrapfigure}[15]{r}{0.5\linewidth}
    \vspace{-\baselineskip}
    \centering
    \includegraphics[width=\linewidth]{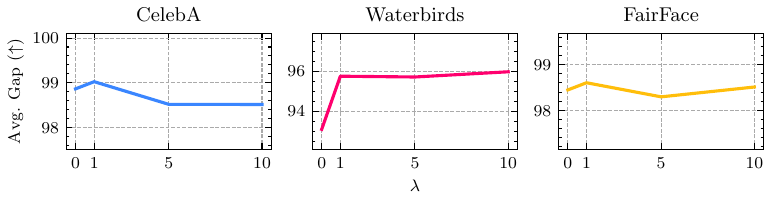}
    \caption{\textbf{Ablating parameter $\lambda$.} \method{} \gap{} when varying parameter $\lambda$ in \celeba{}, \waterbird{}, and \fairface{}. While $\lambda=1$ is optimal in \celeba{} and \fairface{}, \waterbird{} benefits from higher lambdas.}
    \label{fig:lambda}
\end{wrapfigure}
The MIA shows subtle oscillations in \celeba{} and \fairface{} while it drops when adding the \emph{calibration term} in \waterbird{}, getting closer to \retrain{} + \reweight{} value (59.5\% of \method{} \vs{} 53.6\% of \retrain{} + \reweight{}).
Finally, TA and RA are relatively stable across components, with RA showing a negative trend (64.7\% using all three components \vs{} 65.6\% of \pretrain{}).

\cref{fig:lambda}, instead, reports the ablation study on parameter $\lambda$.
We show results for $\lambda\in\{0, 1, 5, 10\}$ and average each experiment over three different seeds.
Overall, all three datasets benefit from the calibration term.
However, while \celeba{} and \fairface{} achieve better results when $\lambda=1$ (\ie, an \gap{} of 99.2 and 98.7), \waterbird{} minimizes the gap when $\lambda=10$ (\ie, an \gap{} of 96.9).
Nonetheless, results are generally stable, and when in doubt, we suggest setting $\lambda=1$ as it always shows an alignment improvement with \retrain{} + \reweight{}.

\begin{table}
    \scriptsize
    \centering
    \caption{\changed{\textbf{Additional Fairness Metrics.} Fairness metrics are computed on each of the three investigated datasets (using the same splits as~\cref{tab:robust_celeba,tab:robust_waterbirds,tab:robust_fairface}). From left to right, we report the method, DP, EP, EO, and WG. \inlinecolorbox{colormethod}{\method{} + \reweight{}} is highlighted.}}
    \vspace{\baselineskip}
    \begin{tabularx}{0.6\linewidth} { 
        L{1.8cm}
        *{4}{Y}
    }
            method & DP & EP & EO & WG \\
        \toprule
            \multicolumn{5}{c}{\textbf{\celeba{}}} \\
            \pretrain{} & 18.7 & 45.3 & 24.2 & 40.6\\
            \retrain{} & 18.9\scriptsize~(0.1) & 51.0\scriptsize~(5.7) & 27.0\scriptsize~(2.9) & 34.4\scriptsize~(-6.1)\\
            \retrain{}+\textsc{rw} & 18.8\scriptsize~(0.1) & 44.6\scriptsize~(-0.7) & 23.9\scriptsize~(-0.3) & 41.3\scriptsize~(0.7)\\
            \cellcolor{colormethod}\method{} & \cellcolor{colormethod}18.8\scriptsize~(0.1) & \cellcolor{colormethod}45.0\scriptsize~(-0.3) & \cellcolor{colormethod}24.0\scriptsize~(-0.1) & \cellcolor{colormethod}41.3\scriptsize~(0.7)\\
        \cmidrule{1-5}
            \multicolumn{5}{c}{\textbf{\waterbird{}}} \\
            \pretrain{} & 20.6 & 36.1 & 26.1 & 56.6\\
            \retrain{} & 23.2\scriptsize~(2.6) & 43.4\scriptsize~(7.3) & 30.4\scriptsize~(4.3) & 49.4\scriptsize~(-7.2)\\
            \retrain{} + \textsc{rw} & 21.5\scriptsize~(0.9) & 40.6\scriptsize~(4.5) & 28.3\scriptsize~(2.2) & 51.6\scriptsize~(-5.0)\\
            \cellcolor{colormethod}\method{} & \cellcolor{colormethod}22.9\scriptsize~(2.3) & \cellcolor{colormethod}38.1\scriptsize~(2.0) & \cellcolor{colormethod}28.3\scriptsize~(2.2) & \cellcolor{colormethod}53.7\scriptsize~(-2.9)\\
        \cmidrule{1-5}
            \multicolumn{5}{c}{\textbf{\fairface{}}} \\
            \pretrain{} & 2.0 & 7.6 & 5.4 & 9.4\\
            \retrain{} & 5.3\scriptsize~(3.3) & 17.9\scriptsize~(10.3) & 9.2\scriptsize~(3.8) & 8.3\scriptsize~(-1.1)\\
            \retrain{} + \textsc{rw} & 3.8\scriptsize~(1.8) & 7.6\scriptsize~(0.0) & 5.6\scriptsize~(0.2) & 6.1\scriptsize~(-3.3)\\
            \cellcolor{colormethod}\method{} & \cellcolor{colormethod}1.1\scriptsize~(-0.9) & \cellcolor{colormethod}8.0\scriptsize~(0.4) & \cellcolor{colormethod}5.8\scriptsize~(0.4) & \cellcolor{colormethod}16.3\scriptsize~(6.9)\\
    \end{tabularx}
    \label{tab:fairness}
\end{table}

\subsection{Additional Fairness Metrics}
\label{appxsub:fairness_metrics}
To further study the method's fairness after unlearning.
We investigate three additional fairness metrics alongside the Equalized Odds (EO) \citep{hardt2016equality}, namely, Demographic Parity \citep{kusner2017counterfactual}, Equal Opportunity \citep{hardt2016equality}, and Worst Group Accuracy (WG) \citep{sagawa2019distributionally,liu2021just}.
To satisfy the Demographic Parity, the model's probability of outputting a positive prediction must be independent of the sensitive attribute.
Therefore, we measure it as: $\text{DP} = |\p(\hat{\targets} = 1 \mid \attributes = 0) - \p(\hat{\targets} = 1 \mid \attributes = 1)|$.
Similarly, to satisfy Equal Opportunity (EP), the model true positive rate must be independent of the sensitive attribute: $\text{EP} = |\p(\hat{\targets} = 1 \mid \targets=1, \attributes=0) - \p(\hat{\targets} = 1 \mid \targets=1, \attributes=1)|$.
Instead, the Worst Group Accuracy measures the average accuracy of the worst group of the test set.
Thus, the Worst Group Accuracy is computed as: $\text{WG} = \underset{g_i}{\min}\,\left\{\frac{1}{|\testds^{g_i}|}\sum_{i=1}^{|\testds^{g_i}|} \mathds{1}\left[ (\model)(\image_i) = \target_i \right]\right\}$,
where $\testds^{g_i}$ is the subset of images of group $g_i$ of the test set.
The lower the first two metrics (DP and EP), the better the model fairness, while the higher the WG, the higher the model robustness.

\Cref{tab:fairness} reports our evaluation using the additional metrics on \celeba{}, \waterbird{}, and \fairface{}.
While plain model retraining generally shows performance degradation on almost all fairness metrics (\eg, +5.7 EP, +2.9 EO, and -6.1 WG, in \celeba{}), using \reweight{} recovers the original model fairness and overall robustness (\ie, -0.7 EP, -0.3 EO, and +0.7 WG, in \celeba{}).
Similarly, the proposed \method{} preserves \pretrain{} robustness by approximating \retrain{} + \reweight{} (\ie, -0.3 EP, -0.1 EO, and +0.7 WG, in \celeba{}).

\end{document}